\documentclass[10pt,twocolumn,letterpaper]{article}
\ifdefined\pdfminorversion
\fi

\PassOptionsToPackage{table}{xcolor}
\usepackage[pagenumbers]{cvpr}
\usepackage{booktabs}
\usepackage{multirow}
\usepackage{graphicx}
\usepackage{algorithm}
\usepackage{algpseudocode}
\usepackage{xcolor}
\definecolor{tedrow}{RGB}{255,248,214}

\usepackage{placeins}  
\usepackage{tabularx}  
\newcolumntype{Y}{>{\centering\arraybackslash}X}
\makeatletter
\renewcommand{\paragraph}{\@startsection{paragraph}{4}{\z@}%
  {3.25ex}{-1em}{\normalfont\normalsize\bfseries}}
\makeatother

\usepackage{pifont}
\newcommand{\cmark}{\ding{51}}\newcommand{\xmark}{\ding{55}}
\newcommand{\R}{\mathbb{R}} \newcommand{\E}{\mathbb{E}}
\newcommand{\cS}{\mathcal{S}} \newcommand{\cD}{\mathcal{D}} \newcommand{\cB}{\mathcal{B}}
\newcommand{\cC}{\mathcal{C}} \newcommand{\cL}{\mathcal{L}} \newcommand{\cG}{\mathcal{G}}
\newcommand{\Lanc}{\mathcal{L}_{\mathrm{anc}}}
\DeclareMathOperator*{\argmin}{arg\,min}

\definecolor{cvprblue}{rgb}{0.21,0.49,0.74}
\usepackage[pagebackref,breaklinks,colorlinks,allcolors=cvprblue,hypertexnames=false]{hyperref}

\hypersetup{
  pdftitle={Learning Text Targets for Generative Multimodal Dataset Distillation},
  pdfauthor={Chongle Ren, Guang Li, Wenbo Huang, Zhenghao Chen, Naoki Saito, Takahiro Ogawa, Miki Haseyama}
}
\title{Learning Text Targets for Generative Multimodal Dataset Distillation}
\author{Chongle Ren$^{1}$\quad
  Guang Li$^{1}$\thanks{Correspondence to Guang Li
    \textless{}guang@lmd.ist.hokudai.ac.jp\textgreater{}.}\quad
  Wenbo Huang$^{2}$\quad Zhenghao Chen$^{3}$\\[2pt]
  Naoki Saito$^{1}$\quad Takahiro Ogawa$^{1}$\quad Miki Haseyama$^{1}$\\[4pt]
  {\normalsize $^{1}$Hokkaido University\quad $^{2}$Southeast University\quad $^{3}$University of Newcastle}}
\date{}

\begin{document}
\maketitle
\begin{abstract}
Multimodal dataset distillation (MDD) aims to reduce the cost of vision--language training by synthesizing a small set of image--text pairs. Generative MDD methods synthesize the images with pretrained generative models and generalize better to unseen architectures than trajectory-matching methods, but they pair each image with a heuristic text embedding. Since the text encoder is frozen, this embedding is the target with which the student learns to align the image, and an embedding that describes the image well is not necessarily a good training target. We propose Text Embedding Distillation (TED), a post-distillation module that learns the text targets of the generated images. To make each text embedding a good training target, TED optimizes it by back-propagating through the entire training of a surrogate student and minimizing the student's loss on real data. Since the surrogate differs from the students used in evaluation, we further introduce an anchor loss that keeps each embedding close to its original, so that the learned targets transfer to unseen architectures. 
Extensive experiments on the Flickr30K and MS-COCO datasets demonstrate that TED consistently improves PDS and EDGE, achieving gains of up to 4.9 points in mean recall.

\end{abstract}
    
\section{Introduction}
\label{sec:intro}

\begin{figure}[t]
\centering
\includegraphics[width=\linewidth]{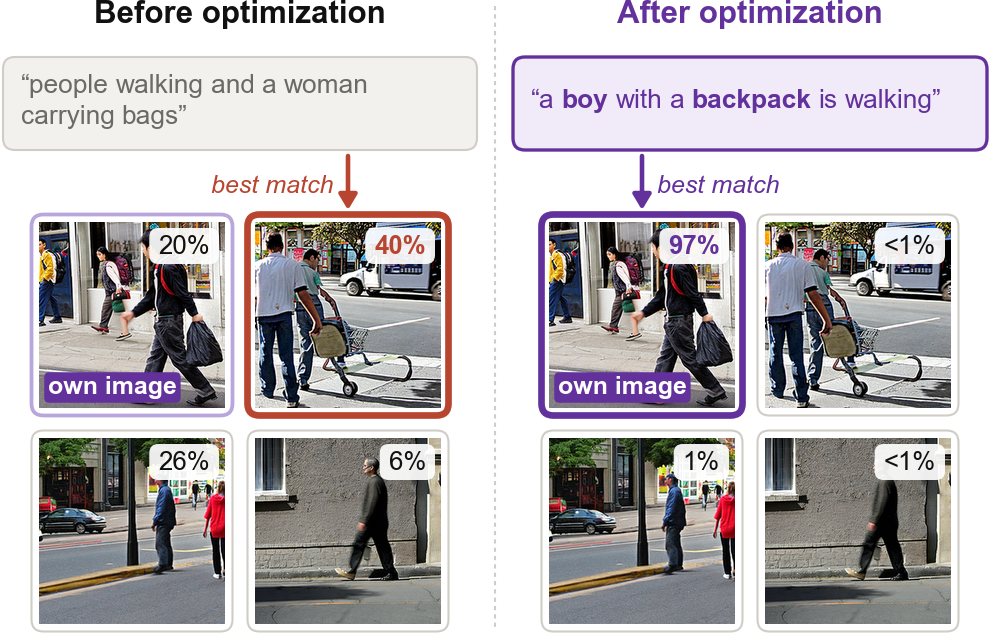}
\caption{\textbf{A text embedding before and after optimization}. The text embedding is paired with a generated image of PDS on Flickr30K (own image, top left); the other three are the generated images it matches best before optimization. Each text is shown by its nearest real caption; percentages give the probability that CLIP retrieves each image for the text among all generated images.}
\label{fig:teaser}
\end{figure}

\indent Vision-language models such as CLIP~\cite{radford2021learning} have become a central paradigm in machine learning, and their success rests on image--text corpora with hundreds of millions of pairs~\cite{schuhmann2022laion,zhang2024vision}. However, training on such large-scale data requires substantial computational resources and training time. This motivates training models on far fewer image--text pairs while retaining most of their performance. \\
\indent Dataset distillation (DD) offers a principled direction by synthesizing a small set of examples that approximates the learning signal of a larger dataset~\cite{li2020soft,li2022awesome,zhao2021datasetcondensation,lee2022dataset,li2026geodm}. Recently, multimodal dataset distillation (MDD)~\cite{wu2024multi,xu2024lors} has extended this idea to multimodal data, condensing a full dataset into a few synthetic pairs. Unlike image-based DD, where each synthetic image is assigned a label from a fixed set of classes, MDD must synthesize the text of each pair, which serves as the supervision for image--text alignment.\\
\indent This setting is typically more challenging than unimodal DD because captions are not restricted to a fixed set of classes, and each synthetic pair must preserve the semantic correspondence between its image and text. Previous MDD methods directly optimize image pixels and text features jointly by matching the training trajectories of experts~\cite{xu2024lors,wu2024multi,zhang2025mdd}. However, this bilevel procedure is expensive: experts must first be trained on the full dataset, and the synthetic data are then updated by repeatedly unrolling segments of their training. Distilling a few hundred pairs may take several GPU days~\cite{guo2026ptmst,jeong2026raha,bo2026apm}. Because the synthetic data are fitted to the training dynamics of one architecture, they generalize poorly to others, and their performance drops sharply when the student uses a different image encoder from the one that produced the trajectories.\\
\indent To mitigate both limitations, recent studies introduce generative MDD. For example, EDGE~\cite{zhao2025edge} and PDS~\cite{choi2026multi} synthesize images with pretrained diffusion models, which avoids bilevel optimization and generalizes better across architectures. EDGE guides the generation with real captions and pairs each image with a real caption and MLLM-written ones. PDS generates each image from a prototype of the real data and pairs it with the mean caption embedding of a cluster. However, this mean captures only what the captions of the cluster share. As a result, neighboring clusters receive similar descriptions, and a text can thus fit another generated image better than its own.
As shown in \cref{fig:teaser} (left), the PDS text of a generated image is more likely to retrieve a neighboring image than its own.\\
\indent This mismatch matters because the text encoder is frozen in multimodal distillation~\cite{xu2024lors,choi2026multi,zhao2025edge}: each text embedding acts as a fixed training target that determines what the student model learns from its paired image. However, semantic alignment and learning effectiveness are not equivalent. An embedding that accurately describes an image may not provide the most effective supervision for a student trained on a compact synthetic set. 
While prior label distillation methods learn supervision targets through student training~\cite{qin2024label,yin2023squeeze,sun2024diversity}, generative MDD introduces an additional challenge, namely, the optimized targets must support downstream learning while preserving transferable cross-modal semantics. The central question is therefore not merely how to better align each generated image with its text, but how to learn text targets that enable a student to generalize effectively to real data. \\
\indent To learn text targets, we propose Text Embedding Distillation (TED), which is a post-distillation module that learns text targets based on their effectiveness in training a student, while keeping the generated images fixed. Specifically, TED optimizes the text embeddings so that a student trained on the synthetic pairs minimizes its loss on real data. Since the student model in multimodal distillation is lightweight, TED back-propagates through its entire training run rather than relying on truncated unrolling or trajectory proxies~\cite{feng2024embarrassingly}. As this student is a surrogate in CLIP space, an anchor loss keeps each text embedding close to its original one, so that the learned targets transfer to students with other image encoders. With TED, the same text retrieves its own image with 97\% probability (\cref{fig:teaser}, right). On Flickr30K and MS-COCO, TED improves PDS and EDGE in every setting by 1.3 to 4.9 points in mean recall. Our contributions are threefold:
\begin{itemize}
    \item We identify the gap between semantic alignment and learning effectiveness in generative MDD, and show that effective text-target updates are image-specific and cannot be obtained from the nearest caption or image features alone.
    \item We propose TED, a simple post-distillation module that learns text targets by unrolling the entire student training process, with an anchor loss that preserves transferable semantic information.
    \item Extensive experiments show that TED consistently improves state-of-the-art generative methods across different settings, and that it remains effective with other evaluation protocols.
\end{itemize}

\section{Related Work}

\begin{figure*}[t]
\centering
\includegraphics[width=\textwidth]{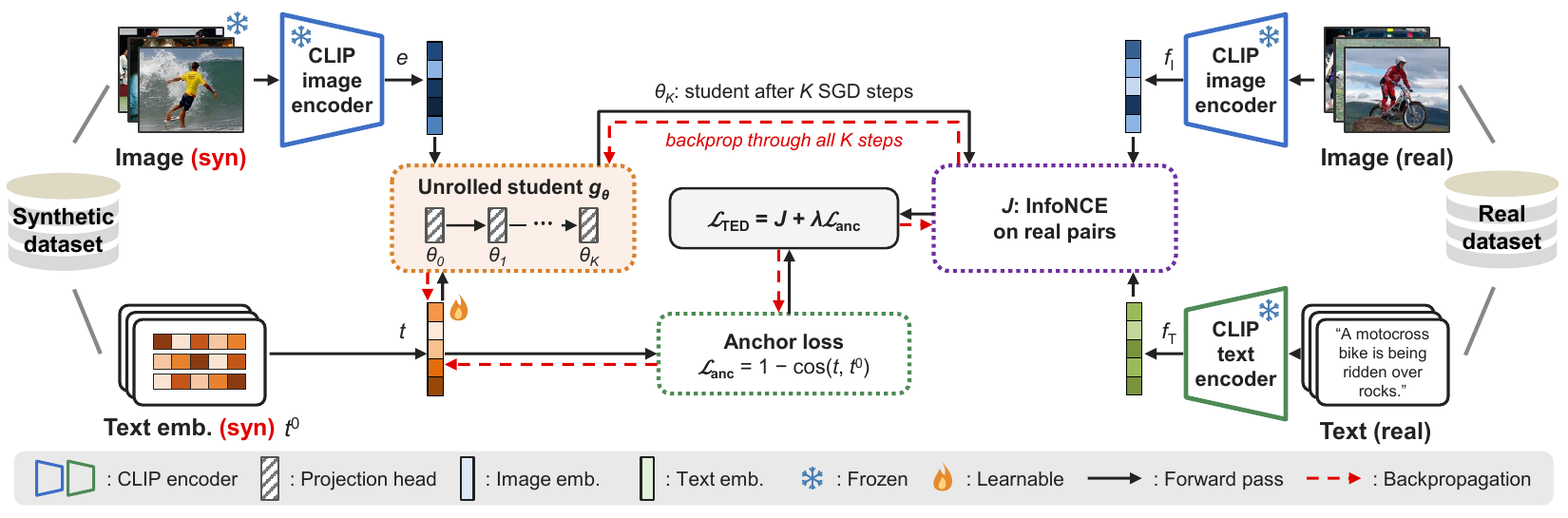}
\caption{\textbf{Overview of TED}.
The synthetic images and the CLIP encoders stay frozen, and only the text embeddings of the synthetic pairs are optimized.
A surrogate student in CLIP space is trained on the synthetic pairs for all SGD steps of the evaluation protocol, and its InfoNCE loss on real pairs is back-propagated through the entire training run to the text embeddings.
An anchor loss keeps each text embedding close to its original one.}
\label{fig:overview}
\end{figure*}

\subsection{Dataset Distillation} 
\indent Dataset distillation synthesizes a small training set on which a model performs close to one trained on the full data~\cite{wang2018datasetdistillation}. It was first posed as a bilevel problem that differentiates the real-data loss through training on the synthetic data. Since unrolling this training is costly, later work matches gradients~\cite{zhao2021datasetcondensation}, parameter trajectories~\cite{cazenavette2022dataset,liu2024att}, or feature distributions~\cite{zhao2023distribution,wang2022cafe,li2026hyperbolic} instead, or solves the inner problem in closed form with a neural tangent kernel, last-layer regression, or linear probing on frozen backbones~\cite{nguyen2021kip,zhou2022frepo,chen2024krr,peng2026closed}. Parameter matching can prune hard-to-match weights~\cite{li2024pruning} or learn their relative importance~\cite{li2024iadd}. Distribution matching can also group examples by difficulty before estimating target statistics~\cite{ma2026difficulty}. Recent methods scale further by decoupling distillation from bilevel optimization~\cite{yin2023squeeze,sun2024diversity}, generating images with pretrained diffusion models~\cite{chan-santiago2025mgd3,su2024d4m,su2024diffusion,gan2026scg,xia2026diver}, or selecting training sets from generated pools~\cite{li2026sas,li2025task}. Diffusion-based approaches have balanced global structure and local detail~\cite{li2024global}, guided sampling with an adaptive memory~\cite{li2025memory}, and used self-supervised features to guide generation~\cite{li2026srg}. A related line of work fixes the inputs and learns supervision targets, such as soft labels or sentence embeddings, through meta-gradients of student training~\cite{bohdal2020flexible,sucholutsky2021soft,deng2022remember}, while RaT-BPTT~\cite{feng2024embarrassingly} reduces the cost through truncated unrolling. Soft labels also provide supervision for distilled gastric images~\cite{li2022soft}. In video distillation, ProtoBlend pairs blended clips with a corresponding mixture of teacher predictions~\cite{ren2026protoblend}. TED builds on this principle but addresses text-target learning for generated image–text pairs, where the optimized targets must remain transferable across image encoders.

\subsection{Multimodal Dataset Distillation} 
\indent In image classification, category-level text and image prototypes can jointly guide synthesis~\cite{zou2025vlcp}, and EVLF aligns the two modalities before the generative backbone~\cite{cai2026evlf}. For image--text retrieval, MTT-VL extends trajectory matching to image--text data, jointly optimizing image pixels and text features~\cite{wu2024multi}, and LoRS additionally learns a low-rank similarity matrix~\cite{xu2024lors}. These methods require expert training and repeated bilevel optimization, and tie the synthetic data to the distillation architecture. MDM matches joint image--text distributions on the unit hypersphere~\cite{jeong2026mdm}, and APM matches the closed-form optimum of a linear projector~\cite{bo2026apm}. Generative methods drop pixel optimization altogether: PDS decodes matched CLIP cluster prototypes with unCLIP~\cite{choi2026multi}, and EDGE fine-tunes a diffusion model with contrastive and diversity losses and adds MLLM-written captions~\cite{zhao2025edge}. Unlike APM~\cite{bo2026apm}, which matches the closed-form optimum of a linear projector, TED directly optimizes the real-data loss along finite student training trajectories, using multiple checkpoints and semantic anchoring. Full unrolling remains affordable because only a lightweight projection head is trained on fixed image features. In PDS and EDGE, the text paired with each image is fixed by a heuristic before any student is trained. TED takes their output and re-solves only the text embeddings. 

\section{Methodology}
\label{sec:method}

\subsection{Preliminaries}
\label{sec:problem}
\indent Let a large-scale image--text dataset $\cD=\{(x_n,y_n)\}_{n=1}^N$ be given, where $x_n$ denotes an image and $y_n$ its paired caption.
The objective of multimodal dataset distillation is to synthesize a compact yet informative dataset $\cS=\{(\tilde x_m,\tilde y_m)\}_{m=1}^M$, $M\ll N$, such that a vision--language model trained on $\cS$ performs comparably to one trained on $\cD$.
An idealized bilevel objective minimizes the real-data loss of the model trained on the synthetic set:\\
\begin{equation}
\begin{aligned}
  \cS^\star &= \argmin_{\cS}\ \E_{\cB\sim\mathcal{P}^{\otimes b}}\big[\cL(\cB;\theta_\cS)\big],\\
  &\quad\text{s.t.}\quad \theta_\cS=\argmin_\theta\ \cL(\cS;\theta),
\end{aligned}
\label{eq:dd}
\end{equation}
where $\mathcal{P}$ is the data distribution and $\cB$ a batch of $b$ real pairs. 
The loss $\cL$ is the symmetric InfoNCE loss~\cite{radford2021learning},
\begin{equation}
  \cL(\cB;\theta)=-\frac{1}{2b}\sum_{i=1}^{b}\Big[\log\frac{e^{s_{ii}/\tau}}{\sum_{j}e^{s_{ij}/\tau}}+\log\frac{e^{s_{ii}/\tau}}{\sum_{j}e^{s_{ji}/\tau}}\Big],
  \label{eq:infonce}
\end{equation}
where $s_{ij}$ is the cosine similarity that the model $\theta$ assigns to the image of the $i$-th pair and the text of the $j$-th pair in $\cB$, and $\tau$ is a temperature.
In practice, the image of each synthetic pair is represented in pixel space, while its text is represented by an embedding $t_m\in\R^d$ in the space of a frozen text encoder $f_T$.
Generative methods synthesize $\tilde x_m$ with a pretrained diffusion model and assign each image a text embedding $t^0_m$ by a heuristic, giving $\cS_\cG=\{(\tilde x_m,t^0_m)\}_{m=1}^M$.
Under their evaluation protocol, the student consists of a frozen, pretrained image encoder $\phi$ and a trainable text projection head $g_\theta$, trained on the synthetic pairs with SGD for $K$ steps~\cite{choi2026multi}.
Since the image features $\phi(\tilde x_m)$ are fixed, $t_m$ determines what the student learns to retrieve for the image of pair $m$, as a label does in classification.

\subsection{Text Embedding Distillation}
\label{sec:ted}

\paragraph{Objective.}
Given $S_G$, TED keeps the generated images fixed and optimizes only the text embeddings $T=[t_1;\ldots;t_M]\in\mathbb{R}^{M\times d}$. Rather than optimizing their direct alignment with the
generated images, we learn these embeddings according to how effectively they train a student to generalize to real data. Specifically, we replace the idealized inner $\arg\min$ in Eq.~(1) with the finite training procedure used in evaluation: the head $g_\theta$ is trained on $\{(\phi(\tilde{x}_m),t_m)\}_{m=1}^{M}$ for $K$ SGD steps from a random initialization $\theta_0$, with checkpoints $\mathcal{C}$. Writing $\theta_c(T;\theta_0)$ for the parameters at checkpoint $c\in\mathcal{C}$, we optimize
\begin{equation}
  \min_{T}\ \E_{\theta_0}\,\E_{\cB}\Big[\frac{1}{|\cC|}\sum_{c\in\cC}\cL\big(\cB;\theta_c(T;\theta_0)\big)\Big],
  \label{eq:ted}
\end{equation}
where $\cB$ is a batch of real training pairs $(\phi(x_n),f_T(y_n))$.
The expectation over $\theta_0$ accounts for student re-initialization; the checkpoint average is a differentiable proxy for the best-checkpoint selection used in evaluation.
Bilevel objectives of this form are usually approximated, because unrolling the inner training is expensive~\cite{cazenavette2022dataset}.
Here, however, the inner problem is a small head trained for about a hundred steps on $M$ pairs, so $\theta_c(T;\theta_0)$ is a cheap, differentiable function of $T$, and \cref{eq:ted} can be optimized by back-propagating through the training run.
The output has the same size and format as $\cS_\cG$; only its text side is replaced.
\paragraph{A surrogate student in CLIP space.}
The evaluation encoder $\phi$ is unknown at distillation time.
We therefore replace $\phi$ by the CLIP image encoder $f_I$, which shares its embedding space with the text encoder $f_T$ and whose training-set features generative methods already compute.
With $e_m=f_I(\tilde x_m)$ and $g_\theta:\R^d\to\R^d$, the surrogate student is trained on $\{(e_m,t_m)\}_{m=1}^M$, and the objective minimized in \cref{eq:ted} becomes
\begin{equation}
\begin{aligned}
  J(T)=\ &\E_{\theta_0}\,\E_{\cB}\Big[\frac{1}{|\cC|}\sum_{c\in\cC}\cL\big(\cB;\theta_c(T;\theta_0)\big)\Big],\\
  &\cB\subset\{(f_I(x_n),f_T(y_n))\}_{n=1}^N,
\end{aligned}
\label{eq:J}
\end{equation}
where each real image in $\cB$ is paired with one of its captions at random.
The surrogate differs from the evaluation students only in the image encoder; the training procedure is identical.

\paragraph{Direction-only parametrization.}
We keep the norm of each initial embedding $t^0_m$ fixed and optimize only
its direction. We parametrize the direction through an unconstrained displacement
$\delta\in\R^{M\times d}$, initialized at zero, with $\delta_m$ its $m$-th row:
\begin{equation}
  t_m(\delta)=\|t^0_m\|\,\frac{u^0_m+\delta_m}{\|u^0_m+\delta_m\|},
  \qquad u^0_m=\frac{t^0_m}{\|t^0_m\|}.
  \label{eq:param}
\end{equation}
At $\delta=0$ this recovers the initial embeddings, and $\delta$ can be
optimized with a standard optimizer. Since every $t_m$ keeps its original
norm, the student sees embeddings on the same scale as before, and any
change in what it learns comes from where they point.
We write $u_m=t_m/\|t_m\|=(u^0_m+\delta_m)/\|u^0_m+\delta_m\|$ for the current direction and $U=[u_1;\dots;u_M]$.


\subsection{Semantic Anchor Regularization}
\label{sec:prior}
Minimizing $J$ alone fits the surrogate student, whose image features come from CLIP, whereas the evaluation students rely on other encoders.
We hypothesize that small embedding changes are more likely to preserve transferable semantic information, whereas larger changes may overfit to surrogate-specific representations.
The upstream embeddings are derived from real captions and thus carry semantics independent of any student.
We therefore regularize each solved direction towards its upstream direction with a cosine anchor loss:
\begin{equation}
  \Lanc(U)=\frac{1}{M}\sum_{m=1}^{M}\big(1-\langle u_m,u^0_m\rangle\big).
  \label{eq:prior}
\end{equation}
Since the norm of each embedding is fixed by \cref{eq:param}, $\Lanc$ measures its entire change.
Similar regularization towards a reference solution is used in transfer learning to preserve pretrained knowledge~\cite{li2018explicit}.

\paragraph{Total objective.} The final objective of TED is
\begin{equation}
  \mathcal{L}_{\text{TED}}(T)=J(T)+\lambda\,\Lanc(U),
  \label{eq:map}
\end{equation}
where $\lambda$ balances distillation and anchoring.
Setting $\lambda=0$ removes the semantic anchor regularization, and $\lambda\to\infty$ keeps the upstream embeddings.

\Cref{fig:overview} illustrates the pipeline of TED, which keeps the generated images fixed and learns only their text embeddings. A surrogate student is trained on the synthetic pairs as in the evaluation protocol, and its loss on real pairs, together with an anchor loss towards the original embeddings, is back-propagated through the entire training to update the text embeddings.

\begin{table*}[t]
\centering
\footnotesize
\setlength{\tabcolsep}{3.9pt}
\setlength{\aboverulesep}{0pt}   
\setlength{\belowrulesep}{0pt}   
\setlength{\extrarowheight}{2pt} 
\caption{Comparison of EDGE and PDS with and without TED on Flickr30K and MS-COCO.  R$_\text{mean}$ (\%) of ResNet and ViT backbones; each ``+ TED'' row keeps the images of the row above and replaces only their text embeddings. ($\uparrow$): gain over the row above.}
\label{tab:main}
\begin{tabular}{ll|ccc|ccc}
\toprule
 & & \multicolumn{3}{c|}{Flickr30K} & \multicolumn{3}{c}{MS-COCO} \\
\midrule
Backbone & Method & $M{=}100$ & $M{=}300$ & $M{=}500$ & $M{=}100$ & $M{=}300$ & $M{=}500$ \\
\midrule
\multirow{4}{*}{ResNet} & EDGE & $19.1\pm0.3$ & $33.2\pm0.1$ & $37.2\pm0.2$ & $8.0\pm0.1$ & $14.4\pm0.1$ & $17.8\pm0.1$ \\
 & \cellcolor[gray]{0.93}EDGE + TED & \cellcolor[gray]{0.93}$24.0\pm0.3$ \textcolor[RGB]{40,150,60}{($\uparrow$\,4.9)} & \cellcolor[gray]{0.93}$36.5\pm0.3$ \textcolor[RGB]{40,150,60}{($\uparrow$\,3.3)} & \cellcolor[gray]{0.93}$41.1\pm0.2$ \textcolor[RGB]{40,150,60}{($\uparrow$\,3.9)} & \cellcolor[gray]{0.93}$10.2\pm0.1$ \textcolor[RGB]{40,150,60}{($\uparrow$\,2.2)} & \cellcolor[gray]{0.93}$17.2\pm0.1$ \textcolor[RGB]{40,150,60}{($\uparrow$\,2.8)} & \cellcolor[gray]{0.93}$19.9\pm0.1$ \textcolor[RGB]{40,150,60}{($\uparrow$\,2.1)} \\
 & PDS & $25.1\pm0.2$ & $38.0\pm0.2$ & $40.8\pm0.1$ & $11.5\pm0.1$ & $16.4\pm0.1$ & $19.2\pm0.1$ \\
 & \cellcolor[gray]{0.93}PDS + TED & \cellcolor[gray]{0.93}$29.5\pm0.2$ \textcolor[RGB]{40,150,60}{($\uparrow$\,4.4)} & \cellcolor[gray]{0.93}$41.6\pm0.2$ \textcolor[RGB]{40,150,60}{($\uparrow$\,3.6)} & \cellcolor[gray]{0.93}$44.7\pm0.2$ \textcolor[RGB]{40,150,60}{($\uparrow$\,3.9)} & \cellcolor[gray]{0.93}$12.8\pm0.0$ \textcolor[RGB]{40,150,60}{($\uparrow$\,1.3)} & \cellcolor[gray]{0.93}$18.1\pm0.1$ \textcolor[RGB]{40,150,60}{($\uparrow$\,1.7)} & \cellcolor[gray]{0.93}$21.0\pm0.1$ \textcolor[RGB]{40,150,60}{($\uparrow$\,1.8)} \\
\midrule
\multirow{4}{*}{ViT} & EDGE & $12.9\pm0.2$ & $21.8\pm0.2$ & $25.8\pm0.1$ & $4.6\pm0.1$ & $10.0\pm0.1$ & $11.9\pm0.2$ \\
 & \cellcolor[gray]{0.93}EDGE + TED & \cellcolor[gray]{0.93}$16.0\pm0.2$ \textcolor[RGB]{40,150,60}{($\uparrow$\,3.1)} & \cellcolor[gray]{0.93}$24.5\pm0.2$ \textcolor[RGB]{40,150,60}{($\uparrow$\,2.7)} & \cellcolor[gray]{0.93}$28.9\pm0.2$ \textcolor[RGB]{40,150,60}{($\uparrow$\,3.1)} & \cellcolor[gray]{0.93}$6.5\pm0.1$ \textcolor[RGB]{40,150,60}{($\uparrow$\,1.9)} & \cellcolor[gray]{0.93}$11.8\pm0.1$ \textcolor[RGB]{40,150,60}{($\uparrow$\,1.8)} & \cellcolor[gray]{0.93}$13.6\pm0.2$ \textcolor[RGB]{40,150,60}{($\uparrow$\,1.7)} \\
 & PDS & $15.1\pm0.1$ & $24.5\pm0.1$ & $28.6\pm0.2$ & $7.8\pm0.1$ & $11.7\pm0.1$ & $13.9\pm0.1$ \\
 & \cellcolor[gray]{0.93}PDS + TED & \cellcolor[gray]{0.93}$19.3\pm0.1$ \textcolor[RGB]{40,150,60}{($\uparrow$\,4.2)} & \cellcolor[gray]{0.93}$27.6\pm0.2$ \textcolor[RGB]{40,150,60}{($\uparrow$\,3.1)} & \cellcolor[gray]{0.93}$31.7\pm0.2$ \textcolor[RGB]{40,150,60}{($\uparrow$\,3.1)} & \cellcolor[gray]{0.93}$9.1\pm0.1$ \textcolor[RGB]{40,150,60}{($\uparrow$\,1.3)} & \cellcolor[gray]{0.93}$13.5\pm0.2$ \textcolor[RGB]{40,150,60}{($\uparrow$\,1.8)} & \cellcolor[gray]{0.93}$16.1\pm0.1$ \textcolor[RGB]{40,150,60}{($\uparrow$\,2.2)} \\
\bottomrule
\end{tabular}
\end{table*}

\section{Experiments}
\label{sec:experiments}

\subsection{Experimental Setup}
\label{sec:exp_setup}

\paragraph{Datasets, Metrics, and Baselines.}
We evaluate on Flickr30K~\cite{young2014image} and MS-COCO~\cite{lin2014microsoft}, where each image is paired with five human-annotated captions.
We focus on the cross-modal retrieval task, which aims to retrieve the top-k relevant samples in the target modality conditioned on a query from the source modality. 
Performance is measured by image-to-text retrieval (TR@$k$) and text-to-image retrieval (IR@$k$) for $k\in\{1,5,10\}$ and by their mean, R$_\text{mean}$. We report retrieval performance on Flickr30k and MS-COCO for budgets $M\in\{100,300,500\}$.
We apply TED to two generative methods, PDS~\cite{choi2026multi} and EDGE~\cite{zhao2025edge}, both run with their public code, so that every upstream set and its TED counterpart share the same images, students and protocol.

\paragraph{Implementation Details.}
\label{par:protocol}
Unless otherwise specified, we use CLIP ViT-L/14~\cite{radford2021learning} as the surrogate and text encoder. The image features of the synthetic and real training images are computed once before the solve.
The displacement $\delta$ is initialized at zero and optimized with Adam for $300$ outer steps. We follow the evaluation protocol of PDS~\cite{choi2026multi} on NF-ResNet-50~\cite{brock2021characterizing} and ViT-Ti/16~\cite{DBLP:conf/iclr/DosovitskiyB0WZ21}. Each evaluation backbone serves as the image encoder of the student, and a trainable linear projection is appended to the frozen text encoder. See the supplementary material for additional implementation details.

\subsection{Main Results}
\label{sec:main}

\paragraph{Improvement on Generative Methods.}
As shown in \cref{tab:main}, TED consistently improves both generative methods on both datasets, both architectures, and all three budgets. Each pair of rows shares images, students, and protocol, so the gains come from the text alone: on Flickr30K with 100 pairs, TED raises PDS from $25.1$ to $29.5$ on ResNet and from $15.1$ to $19.3$ on ViT. The relative gain is largest at the smallest budget, where each text embedding stands for a larger part of the training set, and the weaker upstream method gains more in relative terms in nine of the twelve settings, so the gain grows with how much the heuristic text leaves to correct. On MS-COCO, the absolute gains are smaller, as are the scores themselves, but the relative gains are comparable or larger. TED does not differentiate through either evaluation backbone, yet both improve by similar margins, so the text solved in CLIP space transfers to image encoders through which the solver never back-propagates. These results show that the text side limits what a generated set teaches, whichever generator produced it.

\paragraph{Comparison with Coreset Methods.}
As shown in \cref{tab:coreset}, we compare TED with the coreset methods, which select $100$ real pairs from Flickr30K and train a ResNet student. The coreset methods include K-center~\cite{farahani2009facility}, Herding~\cite{welling2009herding}, the CLIP-score, LAION and DataComp~\cite{gadre2023datacomp}.
Every coreset method stays below both generative methods on all six metrics, and with TED the strongest generated set reaches about twice the IR@10 of Herding, the best coreset method.
TED widens the IR@10 margin over Herding for both generators, from $9.0$ to $15.2$ points for EDGE and from $16.6$ to $22.0$ for PDS.

\paragraph{Comparison with Optimization-based Methods.}
As shown in \cref{tab:sota}, TED improves every retrieval metric of both generative methods across all evaluated settings, enabling them to outperform TESLA-VL and LoRS on all metrics. On Flickr30K with 300 pairs and ResNet, PDS + TED reaches $76\%$ of the full-data student's IR@10 using about $1\%$ of the training images. The advantage is particularly pronounced on ViT, which differs from the NFNet used to distill the trajectory-matching baselines: their IR@1 drops by at least half from ResNet to ViT, whereas PDS + TED loses at most a third. This suggests that TED-enhanced generative datasets transfer more effectively across evaluation architectures.

\begin{table}[t]
\centering
\footnotesize
\setlength{\tabcolsep}{2.5pt}
\setlength{\aboverulesep}{0pt}
\setlength{\belowrulesep}{0pt}
\renewcommand{\arraystretch}{1.15}
\caption{Comparison with coreset methods on Flickr30K. Recall@$k$ (\%) of the ResNet student with 100 pairs, mean$\pm$std over five student initializations. $^\dagger$: reported in~\citet{choi2026multi}.}
\label{tab:coreset}
\begin{tabularx}{\columnwidth}{l|YYYYYY}
\toprule
& \multicolumn{3}{c}{IR@$k$} & \multicolumn{3}{c}{TR@$k$} \\
\midrule
Method & 1 & 5 & 10 & 1 & 5 & 10 \\
\midrule
K-center$^\dagger$ & 2.9{\tiny$\pm$0.1} & 10.4{\tiny$\pm$0.3} & 16.8{\tiny$\pm$0.3} & 5.3{\tiny$\pm$0.3} & 15.6{\tiny$\pm$0.5} & 24.1{\tiny$\pm$0.4} \\
Herding$^\dagger$ & 3.6{\tiny$\pm$0.2} & 12.6{\tiny$\pm$0.5} & 20.1{\tiny$\pm$0.6} & 6.7{\tiny$\pm$0.3} & 19.4{\tiny$\pm$0.7} & 28.2{\tiny$\pm$0.6} \\
CLIP-score$^\dagger$ & 2.5{\tiny$\pm$0.1} & 8.7{\tiny$\pm$0.3} & 14.5{\tiny$\pm$0.3} & 4.7{\tiny$\pm$0.4} & 12.6{\tiny$\pm$0.7} & 18.9{\tiny$\pm$0.7} \\
LAION$^\dagger$ & 2.4{\tiny$\pm$0.1} & 8.5{\tiny$\pm$0.2} & 14.5{\tiny$\pm$0.3} & 4.7{\tiny$\pm$0.3} & 12.4{\tiny$\pm$0.6} & 19.3{\tiny$\pm$0.4} \\
Image-based$^\dagger$ & 2.2{\tiny$\pm$0.2} & 8.6{\tiny$\pm$0.3} & 13.6{\tiny$\pm$0.4} & 4.0{\tiny$\pm$0.3} & 10.5{\tiny$\pm$0.4} & 16.2{\tiny$\pm$0.6} \\
\midrule
EDGE & 6.0{\tiny$\pm$0.2} & 19.6{\tiny$\pm$0.2} & 29.1{\tiny$\pm$0.3} & 7.6{\tiny$\pm$0.6} & 21.3{\tiny$\pm$0.9} & 30.9{\tiny$\pm$0.4} \\
\cellcolor[gray]{0.93}\quad + TED & \cellcolor[gray]{0.93}7.9{\tiny$\pm$0.2} & \cellcolor[gray]{0.93}24.3{\tiny$\pm$0.2} & \cellcolor[gray]{0.93}35.3{\tiny$\pm$0.3} & \cellcolor[gray]{0.93}11.3{\tiny$\pm$0.5} & \cellcolor[gray]{0.93}27.1{\tiny$\pm$0.6} & \cellcolor[gray]{0.93}38.0{\tiny$\pm$0.6} \\
PDS & 8.4{\tiny$\pm$0.1} & 25.6{\tiny$\pm$0.3} & 36.7{\tiny$\pm$0.2} & 11.6{\tiny$\pm$0.4} & 29.1{\tiny$\pm$0.4} & 39.3{\tiny$\pm$0.6} \\
\cellcolor[gray]{0.93}\quad + TED & \cellcolor[gray]{0.93}\textbf{10.2}{\tiny$\pm$0.1} & \cellcolor[gray]{0.93}\textbf{29.7}{\tiny$\pm$0.2} & \cellcolor[gray]{0.93}\textbf{42.1}{\tiny$\pm$0.3} & \cellcolor[gray]{0.93}\textbf{14.5}{\tiny$\pm$0.5} & \cellcolor[gray]{0.93}\textbf{34.3}{\tiny$\pm$0.3} & \cellcolor[gray]{0.93}\textbf{45.9}{\tiny$\pm$0.4} \\
\bottomrule
\end{tabularx}
\end{table}

\begin{table*}[t]
\centering
\footnotesize
\setlength{\tabcolsep}{3pt}
\setlength{\aboverulesep}{0pt}
\setlength{\belowrulesep}{0pt}
\renewcommand{\arraystretch}{1.15}
\caption{Comparison with the state-of-the-art methods on Flickr30K and MS-COCO. Full dataset: the student trained on all training pairs. $^\dagger$: reported in~\citet{choi2026multi}. Best and second-best results in each block are in \textbf{bold} and \underline{underlined}.}
\label{tab:sota}
\begin{tabularx}{\textwidth}{cl|YYYYYY|YYYYYY}
\toprule
 & & \multicolumn{6}{c|}{Flickr30K} & \multicolumn{6}{c}{MS-COCO} \\
\cmidrule{3-14}
$M$ & Method & IR@1 & IR@5 & IR@10 & TR@1 & TR@5 & TR@10 & IR@1 & IR@5 & IR@10 & TR@1 & TR@5 & TR@10 \\
\midrule
\multicolumn{14}{c}{\textit{ResNet}} \\
\midrule
\multirow{6}{*}{100} & TESLA-VL$^\dagger$ & 4.1{\tiny$\pm$0.3} & 14.7{\tiny$\pm$0.9} & 22.9{\tiny$\pm$1.2} & 6.5{\tiny$\pm$0.4} & 17.8{\tiny$\pm$1.4} & 27.3{\tiny$\pm$1.4} & 1.4{\tiny$\pm$0.1} & 5.8{\tiny$\pm$0.2} & 10.2{\tiny$\pm$0.4} & 2.1{\tiny$\pm$0.3} & 7.6{\tiny$\pm$0.3} & 12.5{\tiny$\pm$0.3} \\
 & LoRS$^\dagger$ & 6.3{\tiny$\pm$0.1} & 18.6{\tiny$\pm$0.1} & 28.0{\tiny$\pm$0.2} & 9.1{\tiny$\pm$0.2} & 24.3{\tiny$\pm$0.4} & 34.5{\tiny$\pm$0.8} & 1.8{\tiny$\pm$0.1} & 6.8{\tiny$\pm$0.2} & 11.4{\tiny$\pm$0.4} & 2.3{\tiny$\pm$0.2} & 7.8{\tiny$\pm$0.5} & 12.6{\tiny$\pm$0.4} \\
\cmidrule(lr){2-14}
 & EDGE & 6.0{\tiny$\pm$0.2} & 19.6{\tiny$\pm$0.2} & 29.1{\tiny$\pm$0.3} & 7.6{\tiny$\pm$0.6} & 21.3{\tiny$\pm$0.9} & 30.9{\tiny$\pm$0.4} & 2.1{\tiny$\pm$0.0} & 7.4{\tiny$\pm$0.1} & 12.4{\tiny$\pm$0.1} & 2.7{\tiny$\pm$0.2} & 8.8{\tiny$\pm$0.3} & 14.5{\tiny$\pm$0.3} \\
 & \cellcolor[gray]{0.93}\quad + TED & \cellcolor[gray]{0.93}7.9{\tiny$\pm$0.2} & \cellcolor[gray]{0.93}24.3{\tiny$\pm$0.2} & \cellcolor[gray]{0.93}35.3{\tiny$\pm$0.3} & \cellcolor[gray]{0.93}11.3{\tiny$\pm$0.5} & \cellcolor[gray]{0.93}27.1{\tiny$\pm$0.6} & \cellcolor[gray]{0.93}38.0{\tiny$\pm$0.6} & \cellcolor[gray]{0.93}2.4{\tiny$\pm$0.0} & \cellcolor[gray]{0.93}8.3{\tiny$\pm$0.0} & \cellcolor[gray]{0.93}14.2{\tiny$\pm$0.1} & \cellcolor[gray]{0.93}3.7{\tiny$\pm$0.2} & \cellcolor[gray]{0.93}12.6{\tiny$\pm$0.2} & \cellcolor[gray]{0.93}19.7{\tiny$\pm$0.2} \\
\cmidrule(lr){2-14}
 & PDS & \underline{8.4}{\tiny$\pm$0.1} & \underline{25.6}{\tiny$\pm$0.3} & \underline{36.7}{\tiny$\pm$0.2} & \underline{11.6}{\tiny$\pm$0.4} & \underline{29.1}{\tiny$\pm$0.4} & \underline{39.3}{\tiny$\pm$0.6} & \underline{2.9}{\tiny$\pm$0.0} & \underline{10.5}{\tiny$\pm$0.1} & \underline{17.7}{\tiny$\pm$0.1} & \underline{3.9}{\tiny$\pm$0.2} & \underline{13.3}{\tiny$\pm$0.3} & \underline{21.0}{\tiny$\pm$0.3} \\
 & \cellcolor[gray]{0.93}\quad + TED & \cellcolor[gray]{0.93}\textbf{10.2}{\tiny$\pm$0.1} & \cellcolor[gray]{0.93}\textbf{29.7}{\tiny$\pm$0.2} & \cellcolor[gray]{0.93}\textbf{42.1}{\tiny$\pm$0.3} & \cellcolor[gray]{0.93}\textbf{14.5}{\tiny$\pm$0.5} & \cellcolor[gray]{0.93}\textbf{34.3}{\tiny$\pm$0.3} & \cellcolor[gray]{0.93}\textbf{45.9}{\tiny$\pm$0.4} & \cellcolor[gray]{0.93}\textbf{3.1}{\tiny$\pm$0.1} & \cellcolor[gray]{0.93}\textbf{11.3}{\tiny$\pm$0.1} & \cellcolor[gray]{0.93}\textbf{18.7}{\tiny$\pm$0.1} & \cellcolor[gray]{0.93}\textbf{4.9}{\tiny$\pm$0.2} & \cellcolor[gray]{0.93}\textbf{15.0}{\tiny$\pm$0.2} & \cellcolor[gray]{0.93}\textbf{23.6}{\tiny$\pm$0.2} \\
\cmidrule(lr){1-14}
\multirow{6}{*}{300} & TESLA-VL$^\dagger$ & 10.3{\tiny$\pm$0.3} & 28.8{\tiny$\pm$0.6} & 40.6{\tiny$\pm$0.8} & 14.9{\tiny$\pm$0.9} & 36.2{\tiny$\pm$1.1} & 48.8{\tiny$\pm$1.7} & 3.0{\tiny$\pm$0.1} & 10.7{\tiny$\pm$0.3} & 17.6{\tiny$\pm$0.4} & 5.7{\tiny$\pm$0.2} & 17.0{\tiny$\pm$0.3} & 25.6{\tiny$\pm$0.4} \\
 & LoRS$^\dagger$ & 8.6{\tiny$\pm$0.9} & 23.5{\tiny$\pm$1.5} & 33.5{\tiny$\pm$1.8} & 14.7{\tiny$\pm$1.2} & 33.5{\tiny$\pm$2.8} & 44.1{\tiny$\pm$3.2} & 2.5{\tiny$\pm$0.2} & 8.5{\tiny$\pm$0.4} & 13.8{\tiny$\pm$0.6} & 3.2{\tiny$\pm$0.2} & 10.1{\tiny$\pm$0.4} & 15.9{\tiny$\pm$0.6} \\
\cmidrule(lr){2-14}
 & EDGE & 12.7{\tiny$\pm$0.2} & 33.9{\tiny$\pm$0.3} & 46.2{\tiny$\pm$0.4} & 15.8{\tiny$\pm$0.5} & 38.9{\tiny$\pm$0.6} & 51.5{\tiny$\pm$0.6} & 3.9{\tiny$\pm$0.1} & 12.8{\tiny$\pm$0.1} & 20.6{\tiny$\pm$0.2} & 5.8{\tiny$\pm$0.2} & 17.4{\tiny$\pm$0.1} & 26.1{\tiny$\pm$0.3} \\
 & \cellcolor[gray]{0.93}\quad + TED & \cellcolor[gray]{0.93}14.3{\tiny$\pm$0.2} & \cellcolor[gray]{0.93}36.6{\tiny$\pm$0.3} & \cellcolor[gray]{0.93}49.1{\tiny$\pm$0.2} & \cellcolor[gray]{0.93}19.4{\tiny$\pm$0.9} & \cellcolor[gray]{0.93}43.2{\tiny$\pm$0.6} & \cellcolor[gray]{0.93}56.4{\tiny$\pm$0.7} & \cellcolor[gray]{0.93}4.5{\tiny$\pm$0.1} & \cellcolor[gray]{0.93}14.4{\tiny$\pm$0.1} & \cellcolor[gray]{0.93}22.8{\tiny$\pm$0.2} & \cellcolor[gray]{0.93}\underline{7.6}{\tiny$\pm$0.2} & \cellcolor[gray]{0.93}\textbf{22.2}{\tiny$\pm$0.3} & \cellcolor[gray]{0.93}\underline{31.8}{\tiny$\pm$0.4} \\
\cmidrule(lr){2-14}
 & PDS & \underline{14.8}{\tiny$\pm$0.2} & \underline{38.3}{\tiny$\pm$0.2} & \underline{50.4}{\tiny$\pm$0.2} & \underline{22.6}{\tiny$\pm$0.6} & \underline{44.9}{\tiny$\pm$0.5} & \underline{56.9}{\tiny$\pm$0.6} & \underline{4.6}{\tiny$\pm$0.1} & \underline{15.1}{\tiny$\pm$0.1} & \underline{23.9}{\tiny$\pm$0.1} & 6.6{\tiny$\pm$0.1} & 19.2{\tiny$\pm$0.4} & 28.7{\tiny$\pm$0.3} \\
 & \cellcolor[gray]{0.93}\quad + TED & \cellcolor[gray]{0.93}\textbf{16.4}{\tiny$\pm$0.2} & \cellcolor[gray]{0.93}\textbf{41.8}{\tiny$\pm$0.2} & \cellcolor[gray]{0.93}\textbf{54.9}{\tiny$\pm$0.1} & \cellcolor[gray]{0.93}\textbf{24.3}{\tiny$\pm$1.0} & \cellcolor[gray]{0.93}\textbf{49.5}{\tiny$\pm$0.3} & \cellcolor[gray]{0.93}\textbf{62.8}{\tiny$\pm$0.5} & \cellcolor[gray]{0.93}\textbf{5.0}{\tiny$\pm$0.1} & \cellcolor[gray]{0.93}\textbf{15.9}{\tiny$\pm$0.1} & \cellcolor[gray]{0.93}\textbf{25.3}{\tiny$\pm$0.1} & \cellcolor[gray]{0.93}\textbf{7.7}{\tiny$\pm$0.2} & \cellcolor[gray]{0.93}\underline{22.1}{\tiny$\pm$0.3} & \cellcolor[gray]{0.93}\textbf{32.5}{\tiny$\pm$0.3} \\
\cmidrule(lr){1-14}
 & Full dataset & 29.1{\tiny$\pm$0.3} & 60.4{\tiny$\pm$0.2} & 72.3{\tiny$\pm$0.2} & 49.5{\tiny$\pm$0.5} & 78.2{\tiny$\pm$0.3} & 85.8{\tiny$\pm$0.5} & 12.1{\tiny$\pm$0.1} & 32.3{\tiny$\pm$0.1} & 45.6{\tiny$\pm$0.1} & 26.3{\tiny$\pm$0.4} & 53.1{\tiny$\pm$0.2} & 65.7{\tiny$\pm$0.4} \\
\midrule
\multicolumn{14}{c}{\textit{ViT}} \\
\midrule
\multirow{6}{*}{100} & TESLA-VL$^\dagger$ & 2.1{\tiny$\pm$0.3} & 7.8{\tiny$\pm$0.7} & 13.1{\tiny$\pm$1.2} & 2.6{\tiny$\pm$0.6} & 8.7{\tiny$\pm$0.9} & 13.7{\tiny$\pm$1.4} & 0.5{\tiny$\pm$0.2} & 2.1{\tiny$\pm$0.8} & 3.8{\tiny$\pm$1.4} & 0.2{\tiny$\pm$0.1} & 1.0{\tiny$\pm$0.5} & 1.7{\tiny$\pm$0.9} \\
 & LoRS$^\dagger$ & 2.8{\tiny$\pm$0.1} & 9.9{\tiny$\pm$0.4} & 16.1{\tiny$\pm$0.2} & 5.2{\tiny$\pm$0.3} & 13.6{\tiny$\pm$0.5} & 20.5{\tiny$\pm$0.2} & 0.8{\tiny$\pm$0.1} & 3.0{\tiny$\pm$0.2} & 5.4{\tiny$\pm$0.4} & 0.7{\tiny$\pm$0.2} & 2.4{\tiny$\pm$0.3} & 4.1{\tiny$\pm$0.5} \\
\cmidrule(lr){2-14}
 & EDGE & 4.2{\tiny$\pm$0.2} & 13.4{\tiny$\pm$0.3} & 20.7{\tiny$\pm$0.4} & 5.1{\tiny$\pm$0.4} & 13.4{\tiny$\pm$0.6} & 20.7{\tiny$\pm$0.8} & 1.2{\tiny$\pm$0.1} & 4.7{\tiny$\pm$0.3} & 8.4{\tiny$\pm$0.4} & 1.1{\tiny$\pm$0.1} & 4.6{\tiny$\pm$0.3} & 7.9{\tiny$\pm$0.4} \\
 & \cellcolor[gray]{0.93}\quad + TED & \cellcolor[gray]{0.93}4.9{\tiny$\pm$0.1} & \cellcolor[gray]{0.93}16.8{\tiny$\pm$0.3} & \cellcolor[gray]{0.93}25.8{\tiny$\pm$0.3} & \cellcolor[gray]{0.93}\underline{6.4}{\tiny$\pm$0.2} & \cellcolor[gray]{0.93}\underline{17.4}{\tiny$\pm$0.5} & \cellcolor[gray]{0.93}\underline{24.4}{\tiny$\pm$0.4} & \cellcolor[gray]{0.93}1.9{\tiny$\pm$0.0} & \cellcolor[gray]{0.93}6.8{\tiny$\pm$0.1} & \cellcolor[gray]{0.93}11.6{\tiny$\pm$0.2} & \cellcolor[gray]{0.93}1.8{\tiny$\pm$0.1} & \cellcolor[gray]{0.93}6.2{\tiny$\pm$0.3} & \cellcolor[gray]{0.93}10.8{\tiny$\pm$0.2} \\
\cmidrule(lr){2-14}
 & PDS & \underline{6.1}{\tiny$\pm$0.2} & \underline{17.5}{\tiny$\pm$0.4} & \underline{26.0}{\tiny$\pm$0.3} & 4.6{\tiny$\pm$0.2} & 14.7{\tiny$\pm$0.7} & 22.0{\tiny$\pm$0.3} & \underline{2.2}{\tiny$\pm$0.0} & \underline{8.1}{\tiny$\pm$0.1} & \underline{13.8}{\tiny$\pm$0.1} & \underline{2.4}{\tiny$\pm$0.1} & \underline{7.6}{\tiny$\pm$0.1} & \underline{12.7}{\tiny$\pm$0.3} \\
 & \cellcolor[gray]{0.93}\quad + TED & \cellcolor[gray]{0.93}\textbf{7.4}{\tiny$\pm$0.2} & \cellcolor[gray]{0.93}\textbf{21.0}{\tiny$\pm$0.3} & \cellcolor[gray]{0.93}\textbf{31.6}{\tiny$\pm$0.3} & \cellcolor[gray]{0.93}\textbf{7.5}{\tiny$\pm$0.4} & \cellcolor[gray]{0.93}\textbf{20.0}{\tiny$\pm$0.2} & \cellcolor[gray]{0.93}\textbf{28.4}{\tiny$\pm$0.4} & \cellcolor[gray]{0.93}\textbf{2.5}{\tiny$\pm$0.1} & \cellcolor[gray]{0.93}\textbf{9.1}{\tiny$\pm$0.1} & \cellcolor[gray]{0.93}\textbf{15.5}{\tiny$\pm$0.2} & \cellcolor[gray]{0.93}\textbf{2.6}{\tiny$\pm$0.1} & \cellcolor[gray]{0.93}\textbf{9.6}{\tiny$\pm$0.3} & \cellcolor[gray]{0.93}\textbf{15.4}{\tiny$\pm$0.4} \\
\cmidrule(lr){1-14}
\multirow{6}{*}{300} & TESLA-VL$^\dagger$ & 5.1{\tiny$\pm$0.7} & 16.2{\tiny$\pm$1.0} & 24.5{\tiny$\pm$1.1} & 6.1{\tiny$\pm$0.6} & 18.0{\tiny$\pm$1.0} & 27.3{\tiny$\pm$1.4} & 1.5{\tiny$\pm$0.1} & 5.9{\tiny$\pm$0.1} & 10.1{\tiny$\pm$0.3} & 2.2{\tiny$\pm$0.2} & 7.8{\tiny$\pm$0.2} & 12.6{\tiny$\pm$0.4} \\
 & LoRS$^\dagger$ & 4.1{\tiny$\pm$0.5} & 13.1{\tiny$\pm$1.0} & 20.7{\tiny$\pm$1.3} & 6.2{\tiny$\pm$0.6} & 17.1{\tiny$\pm$1.2} & 25.7{\tiny$\pm$1.8} & 1.0{\tiny$\pm$0.3} & 3.7{\tiny$\pm$1.1} & 6.3{\tiny$\pm$1.8} & 1.3{\tiny$\pm$0.4} & 4.6{\tiny$\pm$1.4} & 7.4{\tiny$\pm$2.0} \\
\cmidrule(lr){2-14}
 & EDGE & 7.5{\tiny$\pm$0.2} & 22.8{\tiny$\pm$0.3} & 33.0{\tiny$\pm$0.4} & 9.1{\tiny$\pm$0.6} & 24.5{\tiny$\pm$0.5} & 34.0{\tiny$\pm$0.8} & 3.3{\tiny$\pm$0.1} & 10.7{\tiny$\pm$0.1} & 17.2{\tiny$\pm$0.1} & 2.8{\tiny$\pm$0.1} & 10.0{\tiny$\pm$0.1} & 16.0{\tiny$\pm$0.3} \\
 & \cellcolor[gray]{0.93}\quad + TED & \cellcolor[gray]{0.93}9.2{\tiny$\pm$0.1} & \cellcolor[gray]{0.93}\underline{26.5}{\tiny$\pm$0.2} & \cellcolor[gray]{0.93}\underline{37.4}{\tiny$\pm$0.3} & \cellcolor[gray]{0.93}9.8{\tiny$\pm$0.8} & \cellcolor[gray]{0.93}26.5{\tiny$\pm$0.5} & \cellcolor[gray]{0.93}\underline{37.9}{\tiny$\pm$1.0} & \cellcolor[gray]{0.93}3.5{\tiny$\pm$0.0} & \cellcolor[gray]{0.93}12.1{\tiny$\pm$0.0} & \cellcolor[gray]{0.93}19.5{\tiny$\pm$0.1} & \cellcolor[gray]{0.93}\underline{3.7}{\tiny$\pm$0.1} & \cellcolor[gray]{0.93}\underline{12.6}{\tiny$\pm$0.3} & \cellcolor[gray]{0.93}\underline{19.7}{\tiny$\pm$0.4} \\
\cmidrule(lr){2-14}
 & PDS & \underline{9.6}{\tiny$\pm$0.3} & 25.9{\tiny$\pm$0.5} & 36.9{\tiny$\pm$0.3} & \underline{10.3}{\tiny$\pm$0.5} & \underline{27.0}{\tiny$\pm$0.7} & 37.5{\tiny$\pm$0.8} & \underline{3.7}{\tiny$\pm$0.1} & \underline{12.5}{\tiny$\pm$0.1} & \underline{20.3}{\tiny$\pm$0.1} & 3.4{\tiny$\pm$0.1} & 11.7{\tiny$\pm$0.3} & 18.7{\tiny$\pm$0.3} \\
 & \cellcolor[gray]{0.93}\quad + TED & \cellcolor[gray]{0.93}\textbf{11.1}{\tiny$\pm$0.2} & \cellcolor[gray]{0.93}\textbf{29.3}{\tiny$\pm$0.4} & \cellcolor[gray]{0.93}\textbf{40.6}{\tiny$\pm$0.6} & \cellcolor[gray]{0.93}\textbf{12.8}{\tiny$\pm$0.9} & \cellcolor[gray]{0.93}\textbf{30.6}{\tiny$\pm$0.5} & \cellcolor[gray]{0.93}\textbf{41.4}{\tiny$\pm$0.8} & \cellcolor[gray]{0.93}\textbf{4.1}{\tiny$\pm$0.0} & \cellcolor[gray]{0.93}\textbf{14.0}{\tiny$\pm$0.2} & \cellcolor[gray]{0.93}\textbf{22.2}{\tiny$\pm$0.1} & \cellcolor[gray]{0.93}\textbf{4.5}{\tiny$\pm$0.1} & \cellcolor[gray]{0.93}\textbf{14.4}{\tiny$\pm$0.4} & \cellcolor[gray]{0.93}\textbf{22.1}{\tiny$\pm$0.4} \\
\cmidrule(lr){1-14}
 & Full dataset & 23.5{\tiny$\pm$0.3} & 50.8{\tiny$\pm$0.3} & 63.0{\tiny$\pm$0.3} & 37.5{\tiny$\pm$0.5} & 64.5{\tiny$\pm$0.3} & 74.3{\tiny$\pm$0.5} & 11.2{\tiny$\pm$0.1} & 29.1{\tiny$\pm$0.2} & 40.8{\tiny$\pm$0.1} & 18.7{\tiny$\pm$0.2} & 41.9{\tiny$\pm$0.4} & 54.1{\tiny$\pm$0.4} \\
\bottomrule
\end{tabularx}
\end{table*}

\begin{figure*}[t]
\centering
\includegraphics[width=\textwidth]{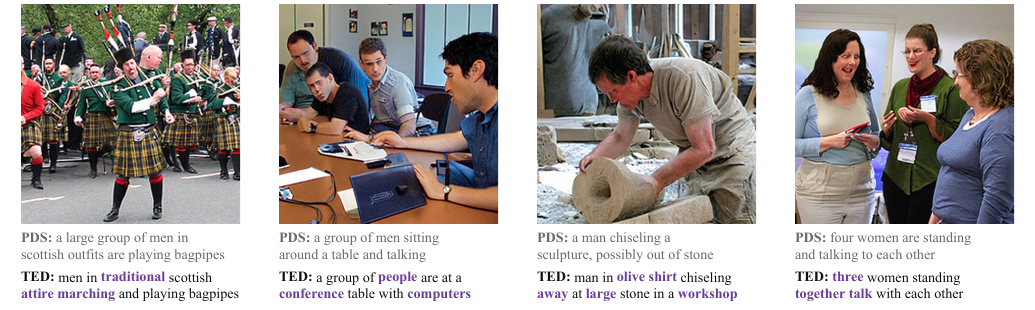}
\caption{PDS and TED text embeddings read as captions. Under each synthetic image: the nearest real training caption of the PDS embedding $t^0$ (gray) and of the TED embedding $t^\star$ (black). Highlighted: words absent from the PDS caption.}
\label{fig:text_change}
\end{figure*}
\paragraph{Qualitative Analysis.}
As shown in \cref{fig:text_change}, since TED outputs an embedding rather than a sentence, we interpret each solved embedding by its nearest real training caption.
TED moves each embedding by a small angle, and for 190 of the 300 images the nearest caption stays the same.
Where it changes, TED keeps the scene of the PDS text and adds or corrects what the generated image shows: the pipers marching in traditional attire, the computers on the conference table, the olive shirt and the workshop of the man chiseling stone, and three women instead of four.
These details belong to the individual image rather than to the cluster of captions that the PDS text summarizes, and they separate each pair from its neighbors.

\subsection{Ablation Study}
\label{sec:ablation}

\begin{table}[t]
\centering
\footnotesize
\caption{Ablation study on the objective of TED. First $K$: only the first $K$ SGD steps are unrolled; Full: the whole run. $\Delta$: gain over PDS, mean of the two architectures.}
\label{tab:abl_components}
\setlength{\tabcolsep}{3pt}
\setlength{\aboverulesep}{0pt}
\setlength{\belowrulesep}{0pt}
\setlength{\extrarowheight}{2pt}
\begin{tabularx}{\columnwidth}{lc|YYY}
\toprule
Unrolled steps & $\Lanc$ & ResNet & ViT & $\Delta$ \\
\midrule
PDS & -- & 38.0$\pm$0.2 & 24.5$\pm$0.1 & -- \\
\midrule
First $K{=}1$ & \cmark & 38.1$\pm$0.3 & 24.8$\pm$0.1 & $+0.2$ \\
First $K{=}3$ & \cmark & 38.8$\pm$0.3 & 25.2$\pm$0.3 & $+0.8$ \\
First $K{=}10$ & \cmark & 40.6$\pm$0.3 & 26.7$\pm$0.1 & $+2.4$ \\
\midrule
Full & \xmark & 38.3$\pm$0.1 & 26.9$\pm$0.1 & $+1.3$ \\
\rowcolor[gray]{0.93}Full (TED) & \cmark & \textbf{41.6}$\pm$0.2 & \textbf{27.6}$\pm$0.2 & $\mathbf{+3.4}$ \\
\bottomrule
\end{tabularx}
\end{table}

\paragraph{Unroll Depth.}
As shown in \cref{tab:abl_components}, the gain of TED grows with the number of unrolled steps $K$ and, averaged over the two architectures, is largest when the solver unrolls the entire training run.
Unrolling a single step barely changes the student, raising the mean R$_\text{mean}$ only from $31.3$ to $31.5$, and the gain rises steeply from three to ten steps.
The text that most reduces the loss after the first update is thus not the text that best shapes the student at the end of training, where it is evaluated.
Because the student is a projection head trained on a small number of synthetic pairs, unrolling the whole run remains cheap and removes the need to choose a truncation length.

\paragraph{Anchor Loss.}
As shown in \cref{tab:abl_components}, unrolling the whole run without $\Lanc$ improves ViT considerably but ResNet only marginally, whereas adding $\Lanc$ brings a large gain on ResNet and a further gain on ViT, so that the benefit of the full unroll transfers to both architectures.
We attribute the gap between the two architectures to surrogate overfitting, since without an anchor the optimization can exploit regularities of the CLIP space that the ViT student tolerates better than the ResNet student.
Both components are therefore necessary, with the unrolled objective providing the direction in which each text embedding moves and the anchor limiting how far it moves so that the change transfers.

\begin{table}[t]
\centering
\footnotesize
\caption{Analysis of the anchor weight $\lambda$ on the validation split. Surrogate loss: InfoNCE on held-out real pairs in CLIP space.}

\label{tab:lambda}
\setlength{\tabcolsep}{3pt}
\setlength{\aboverulesep}{0pt}
\setlength{\belowrulesep}{0pt}
\setlength{\extrarowheight}{2pt}
\begin{tabularx}{\columnwidth}{l|YYY|c}
\toprule
$\lambda$ & ResNet & ViT & $\Delta$ & Surrogate loss \\
\midrule
PDS & 36.1$\pm$0.3 & 23.1$\pm$0.1 & -- & 4.73 \\
\midrule
0 & 37.6$\pm$0.3 & 25.6$\pm$0.2 & $+2.0$ & 4.01 \\
1 & 39.6$\pm$0.1 & \textbf{26.5}$\pm$0.1 & $+3.5$ & 4.08 \\
\rowcolor[gray]{0.93}3 & \textbf{40.1}$\pm$0.3 & \textbf{26.5}$\pm$0.3 & $\mathbf{+3.7}$ & 4.17 \\
10 & 39.6$\pm$0.2 & 25.5$\pm$0.2 & $+3.0$ & 4.32 \\
\bottomrule
\end{tabularx}
\end{table}

\paragraph{Anchor Weight.}
As shown in \cref{tab:lambda}, the held-out surrogate loss falls monotonically as $\lambda$ decreases, whereas the validation gain peaks at $\lambda{=}3$ both for ResNet and for the mean over the two architectures.
On ResNet, the validation R$_\text{mean}$ rises from $37.6$ without an anchor to $40.1$ at $\lambda{=}3$ before falling back to $39.6$ at $\lambda{=}10$, compared with $36.1$ for PDS.
The surrogate loss, which always favors $\lambda\to0$, therefore cannot be used to select $\lambda$, while a single value chosen on validation data serves both architectures.

\begin{figure}[t]
\centering
\includegraphics[width=\linewidth]{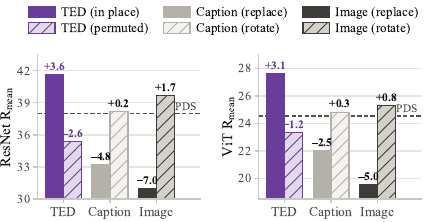}
\caption{Analysis of alternative changes to the text embeddings of the PDS pairs. Colors denote the direction of the change. Solid bars: TED, or $t^0$ replaced by the target; hatched bars: $t^0$ rotated by the angle of TED, along the TED move of another image or towards the target.}
\label{fig:motivation}
\end{figure}

\begin{figure}[t]
\centering
\includegraphics[width=\linewidth]{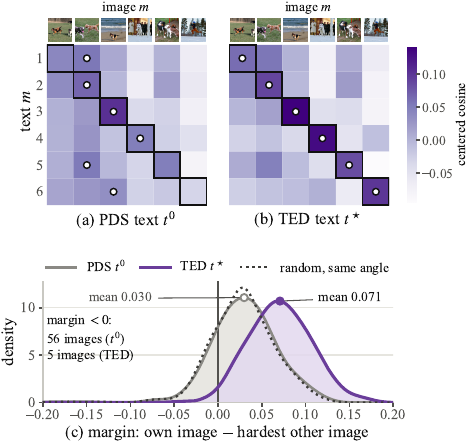}
\caption{Analysis of within-set distinctiveness. (a, b)~Row-centered text--image cosine for six pairs with similar PDS texts; outline: matched pair; dot: best image of each text. (c)~Margin of each text: cosine to its own image minus the highest cosine to another image. Gray dotted: $t^0$ rotated by the angle of TED in a random direction.}
\label{fig:distinct}
\end{figure}

\subsection{Analysis}
\label{sec:analysis}

\label{sec:objective}

\begin{table}[t]
\centering
\footnotesize
\caption{Comparison of objectives for the text embeddings on Flickr30K. All objectives share the generated images, the parametrization, and the anchor loss. We report $R_{\mathrm{mean}}$ and the image--text alignment margin within the distilled set.}
\label{tab:objective}
\setlength{\tabcolsep}{3pt}
\setlength{\aboverulesep}{0pt}
\setlength{\belowrulesep}{0pt}
\setlength{\extrarowheight}{2pt}
\begin{tabularx}{\columnwidth}{l|YYc|c}
\toprule
Objective & ResNet & ViT & $\Delta$ & Margin \\
\midrule
PDS & 38.0$\pm$0.2 & 24.5$\pm$0.1 & -- & 0.030 \\
Direct alignment & 40.0$\pm$0.4 & 25.7$\pm$0.3 & $+1.6$ & 0.267 \\
Trajectory matching & 40.3$\pm$0.2 & 26.6$\pm$0.2 & $+2.2$ & 0.060 \\
APM-style matching~\cite{bo2026apm} & 37.3$\pm$0.2 & 24.8$\pm$0.1 & $-0.2$ & 0.062 \\
\rowcolor[gray]{0.93}TED & \textbf{41.6}$\pm$0.2 & \textbf{27.6}$\pm$0.2 & $\mathbf{+3.4}$ & 0.071 \\
\bottomrule
\end{tabularx}
\end{table}

\paragraph{Choice of Objective.} To examine whether stronger image--text alignment or alternative matching objectives yield equally effective text targets, we compare TED with direct alignment, trajectory matching, and APM-style closed-form matching~\cite{bo2026apm} under a controlled setup. Direct alignment minimizes the InfoNCE loss between the generated images and their text embeddings. All methods share the same generated images, surrogate, parametrization, and anchor loss. As shown in Tab.~\ref{tab:objective}, direct alignment achieves a much larger within-set margin than TED, yet performs worse on both architectures. Trajectory matching improves over PDS but remains below TED, while APM-style matching provides no consistent improvement despite achieving a margin comparable to TED. These results indicate that the choice of learning objective, rather than semantic anchoring alone, is critical to the effectiveness of distilled text targets.

\paragraph{Analysis of Text Embeddings.} We examine whether TED's gains can be reproduced by simpler, image-aligned targets or arbitrary embedding updates. Keeping the PDS images fixed, we replace each text embedding with its nearest real caption or corresponding CLIP image feature, or rotate it towards these targets by the average update angle of TED ($17^\circ$). We also permute TED's updates across images while preserving their magnitudes. As shown in \cref{fig:motivation}, replacing the text with its nearest caption reduces ResNet R$_\text{mean}$ by $4.8$ points, while permuting TED's updates reduces it by $2.6$ points. Only TED's image-specific updates consistently improve both architectures, indicating that the direction of each learned target matters beyond semantic alignment or perturbation magnitude.

\paragraph{Within-Set Distinctiveness.}
As shown in \cref{fig:distinct}, TED makes each synthetic text identify its own image among all generated images, which the PDS text often fails to do because it describes a cluster of real captions and neighboring clusters receive similar descriptions.
The six dog images in \cref{fig:distinct}(a) illustrate this, since several PDS texts match another image better than their own, while every TED text matches its own image best.
Over the whole set, the margin between each text's own image and its hardest other image becomes clearly larger after TED and almost no text still prefers another image, which shows that the gain comes from the direction that TED chooses rather than from the size of the move.
TED thus turns each embedding towards its own image and away from its closest competitor, so that each pair gives the student an unambiguous target.

\paragraph{Generalization to the Trajectory-Matching Setting.}

\begin{table}[t]
\centering
\footnotesize
\caption{Comparison under the training protocol and in the text space of trajectory-matching methods. R$_\text{mean}$ (\%) in the BERT space of LoRS, mean $\pm$ std over five student initializations.}
\label{tab:beyond}
\setlength{\tabcolsep}{3pt}
\setlength{\aboverulesep}{0pt}
\setlength{\belowrulesep}{0pt}
\setlength{\extrarowheight}{2pt}
\begin{tabularx}{\columnwidth}{c|l|YY}
\toprule
$M$ & Method & ResNet & ViT \\
\midrule
\multirow{4}{*}{100} & LoRS & $7.4\pm0.1$ & $4.6\pm0.0$ \\
 & \cellcolor[gray]{0.93}LoRS + TED & \cellcolor[gray]{0.93}$13.3\pm0.2$ \textcolor[RGB]{40,150,60}{($\uparrow$\,5.9)} & \cellcolor[gray]{0.93}$8.5\pm0.1$ \textcolor[RGB]{40,150,60}{($\uparrow$\,3.8)} \\
& PDS & $9.2\pm0.2$ & $2.6\pm1.2$ \\
 & \cellcolor[gray]{0.93}PDS + TED & \cellcolor[gray]{0.93}$\mathbf{18.1}\pm0.3$ \textcolor[RGB]{40,150,60}{($\uparrow$\,8.9)} & \cellcolor[gray]{0.93}$\mathbf{12.4}\pm0.3$ \textcolor[RGB]{40,150,60}{($\uparrow$\,9.8)} \\
\midrule
\multirow{4}{*}{500} & LoRS & $13.3\pm0.2$ & $6.8\pm0.2$ \\
 & \cellcolor[gray]{0.93}LoRS + TED & \cellcolor[gray]{0.93}$20.4\pm0.2$ \textcolor[RGB]{40,150,60}{($\uparrow$\,7.1)} & \cellcolor[gray]{0.93}$12.0\pm0.4$ \textcolor[RGB]{40,150,60}{($\uparrow$\,5.2)} \\
 & PDS & $21.2\pm0.1$ & $13.1\pm1.0$ \\
 & \cellcolor[gray]{0.93}PDS + TED & \cellcolor[gray]{0.93}$\mathbf{23.1}\pm0.4$ \textcolor[RGB]{40,150,60}{($\uparrow$\,1.9)} & \cellcolor[gray]{0.93}$\mathbf{15.8}\pm0.4$ \textcolor[RGB]{40,150,60}{($\uparrow$\,2.7)} \\
\bottomrule
\end{tabularx}
\end{table}

As shown in \cref{tab:beyond}, we further evaluate TED in the setting of trajectory-matching methods~\cite{wu2024multi,xu2024lors}, which differs from the PDS protocol in two ways: the student trains its image encoder together with the text head, and the text embeddings lie in a BERT space rather than in the CLIP space.
TED is re-solved in this space with the same solver and anchor weight; its surrogate student maps BERT text onto CLIP image features, and for LoRS it keeps the synthetic images and starts from the learned text.
TED improves both PDS and LoRS at both budgets, although its surrogate never trains an image encoder, and PDS + TED remains the strongest set on both architectures.
LoRS learns which synthetic pairs count as positives, whereas TED changes the text of each pair, so the two are complementary.
These results indicate that the benefit of TED is tied neither to the PDS protocol nor to the CLIP text space.

\paragraph{Efficiency Analysis.}

\begin{table}[t]
\centering
\footnotesize
\caption{Distillation cost. Expert: expert trajectories trained on the full dataset; Generator: a pretrained generative model. $^\dagger$: reported in~\cite{zhao2025edge}; the other times are measured on one NVIDIA RTX A6000.}
\label{tab:cost}
\setlength{\tabcolsep}{5pt}
\setlength{\aboverulesep}{0pt}
\setlength{\belowrulesep}{0pt}
\setlength{\extrarowheight}{2pt}
\begin{tabular}{l|ccc}
\toprule
Method & Expert & Generator & Time \\
\midrule
MTT-VL$^\dagger$~\cite{wu2024multi} & \cmark & \xmark & $\sim$96 h \\
LoRS$^\dagger$~\cite{xu2024lors} & \cmark & \xmark & $\sim$150 h \\
PDS~\cite{choi2026multi} & \xmark & \cmark & $\sim$1.8 h \\
\rowcolor[gray]{0.93}PDS + TED & \xmark & \cmark & $\sim$1.8 h + 0.3 h \\
\bottomrule
\end{tabular}
\end{table}

We compare the computational cost of different methods in \cref{tab:cost}.
Both trajectory-matching methods require expert trajectories on the full dataset, whereas PDS + TED keeps the requirements of PDS, a pretrained generator and no experts.
MTT-VL and LoRS take about $96$ and $150$ hours, whereas TED adds about $0.3$ hours to the $1.8$ hours of PDS, less than a fifth of its generation time.
TED optimizes only the $M$ text embeddings through a projection head on precomputed CLIP features. These results indicate that TED is a lightweight post-processing step for existing synthetic datasets.

\section{Conclusion}
\label{sec:conclusion}

We investigated the role of text targets in generative MDD, highlighting that semantic alignment alone does not necessarily yield effective supervision for student learning. We proposed Text Embedding Distillation (TED), which optimizes the text embeddings of fixed generated images by differentiating through complete surrogate-student training, with semantic anchoring to improve transfer across image encoders. Experiments on Flickr30K and MS-COCO show consistent improvements over PDS and EDGE across budgets and evaluation architectures. Further analyses show that TED's gains are not matched by direct alignment or simple caption-based refinement, and that the method remains effective under additional training protocols. These findings highlight the importance of optimizing text targets for their effect on downstream learning.

{
    \small
    \bibliographystyle{ieeenat_fullname}
    \bibliography{main}
}

\clearpage
\section*{Appendix}
\raggedbottom
\setcounter{topnumber}{4}
\setcounter{bottomnumber}{3}
\setcounter{totalnumber}{6}
\setcounter{dbltopnumber}{4}
\renewcommand{\topfraction}{0.95}
\renewcommand{\bottomfraction}{0.9}
\renewcommand{\textfraction}{0.05}
\renewcommand{\floatpagefraction}{0.65}
\renewcommand{\dbltopfraction}{0.95}
\renewcommand{\dblfloatpagefraction}{0.65}
\setlength{\textfloatsep}{10pt plus 2pt minus 2pt}
\setlength{\dbltextfloatsep}{10pt plus 2pt minus 2pt}
\setlength{\floatsep}{8pt plus 2pt minus 2pt}
\setlength{\dblfloatsep}{8pt plus 2pt minus 2pt}
\makeatletter
\setlength{\@fptop}{0pt}
\setlength{\@dblfptop}{0pt}
\setlength{\@fpsep}{10pt plus 2pt minus 2pt}
\setlength{\@dblfpsep}{10pt plus 2pt minus 2pt}
\makeatother

\setcounter{section}{0}\setcounter{table}{0}\setcounter{figure}{0}\setcounter{equation}{0}
\renewcommand{\thesection}{S\arabic{section}}
\renewcommand{\thetable}{S\arabic{table}}
\renewcommand{\thefigure}{S\arabic{figure}}
\renewcommand{\theequation}{S\arabic{equation}}
\renewcommand{\theHsection}{supp.\arabic{section}}
\renewcommand{\theHtable}{supp.\arabic{table}}
\renewcommand{\theHfigure}{supp.\arabic{figure}}
\renewcommand{\theHequation}{supp.\arabic{equation}}

This supplementary material contains
the derivation of the TED objective (\cref{sec:supp_derivation}),
implementation details (\cref{sec:supp_impl}),
a description of the synthetic sets and text baselines (\cref{sec:supp_sets}),
full results with all retrieval metrics (\cref{sec:supp_full}),
results under the protocol and text space of trajectory-matching methods (\cref{sec:supp_beyond}),
additional ablations (\cref{sec:supp_ablation}),
an analysis of the solved embeddings (\cref{sec:supp_analysis}),
qualitative results (\cref{sec:supp_qualitative})
and a discussion of limitations and broader impact (\cref{sec:supp_limitations}).

\section{Derivation of the TED Objective}
\label{sec:supp_derivation}

\paragraph{From dataset distillation to the TED objective.}
Dataset distillation seeks the synthetic set whose trained model minimizes the real-data loss,
\begin{equation}
  \min_{\cS}\ \E_{\cB}\big[\cL(\cB;\theta_\cS)\big]
  \quad\text{s.t.}\quad \theta_\cS=\argmin_\theta\ \cL(\cS;\theta),
  \label{eq:supp_dd}
\end{equation}
where $\cB$ is a batch of real pairs and $\cL$ is the symmetric InfoNCE loss.
We obtain the TED objective from \cref{eq:supp_dd} in three steps.
First, TED keeps the generated images, so the synthetic set depends only on the text embeddings, $\cS(T)=\{(\tilde x_m,t_m)\}_{m=1}^M$, and the outer variable reduces to $T$.

Second, the inner $\argmin$ is not what the evaluation computes: the student is trained for $K$ SGD steps from a random initialization $\theta_0$.
Let $\Phi_k(\,\cdot\,;T)$ denote the $k$-th step on the synthetic pairs, given in \cref{eq:supp_sgd}.
The parameters after $k$ steps are
\begin{equation}
  \theta_k(T;\theta_0)=\big(\Phi_{k-1}\circ\dots\circ\Phi_0\big)(\theta_0;T),
  \label{eq:supp_unroll}
\end{equation}
where the momentum buffer is carried along with $\theta$.
Replacing $\theta_\cS$ in \cref{eq:supp_dd} by $\theta_K(T;\theta_0)$ and averaging over the initialization gives
\begin{equation}
  \min_T\ \E_{\theta_0}\E_{\cB}\big[\cL\big(\cB;\theta_K(T;\theta_0)\big)\big].
  \label{eq:supp_last}
\end{equation}

Third, the protocol reports the best of the checkpoints $\cC\subset\{1,\dots,K\}$ instead of the last step.
Let $\ell_c(T;\theta_0)=\E_{\cB}\big[\cL(\cB;\theta_c(T;\theta_0))\big]$ be the real-data loss at checkpoint $c$.
The loss of the best checkpoint, $\min_{c}\ell_c$, is not smooth in $T$, and we bound it by the average:
\begin{equation}
  \min_{c\in\cC}\ \ell_c(T;\theta_0)\ \le\ \frac{1}{|\cC|}\sum_{c\in\cC}\ell_c(T;\theta_0).
  \label{eq:supp_bound}
\end{equation}
Taking the expectation of the right-hand side over $\theta_0$ and minimizing over $T$ gives the TED objective,
\begin{equation}
  \min_{T}\ \E_{\theta_0}\,\E_{\cB}\Big[\frac{1}{|\cC|}\sum_{c\in\cC}\cL\big(\cB;\theta_c(T;\theta_0)\big)\Big].
  \label{eq:supp_ted}
\end{equation}
The bound concerns the real-data loss; the checkpoint is selected by its retrieval score, which is not differentiable, so \cref{eq:supp_ted} is a proxy for the reported score and not an equivalent form.
Replacing the evaluation encoder $\phi$ by the CLIP image encoder $f_I$, so that the student is trained on $\{(f_I(\tilde x_m),t_m)\}_{m=1}^M$ and $\cB$ is drawn from $\{(f_I(x_n),f_T(y_n))\}_{n=1}^N$, turns the objective of \cref{eq:supp_ted} into the surrogate objective $J(T)$, whose expectations are estimated by sampling $\theta_0$ and $\cB$ at every outer step.

\paragraph{Gradient through the training run.}
With learning rate $\eta_k$, momentum $\mu$ and weight decay $\omega$, the step $\Phi_k$ is
\begin{equation}
\begin{aligned}
  v_{k+1} &= \mu v_k+\nabla_\theta\cL_\cS(\theta_k;T)+\omega\theta_k,\\
  \theta_{k+1} &= \theta_k-\eta_k v_{k+1},
\end{aligned}
\label{eq:supp_sgd}
\end{equation}
for $k=0,\dots,K-1$, where $v_0=0$ and $\cL_\cS$ is the InfoNCE loss on a batch of synthetic pairs.
We derive the gradient for one sample of $\theta_0$ and $\cB$; the gradient of $J$ is its expectation.
Let $H_k=\nabla^2_\theta\cL_\cS(\theta_k;T)$ and $G_k=\nabla_T\nabla_\theta\cL_\cS(\theta_k;T)$.
Differentiating \cref{eq:supp_sgd} gives
\begin{equation}
\begin{aligned}
  \mathrm{d}v_{k+1} &= \mu\,\mathrm{d}v_k+(H_k+\omega I)\,\mathrm{d}\theta_k+G_k\,\mathrm{d}T,\\
  \mathrm{d}\theta_{k+1} &= \mathrm{d}\theta_k-\eta_k\,\mathrm{d}v_{k+1},
\end{aligned}
\label{eq:supp_diff}
\end{equation}
and, with $r_k=\nabla_\theta\cL(\cB;\theta_k)$, the objective changes by $\mathrm{d}J=\frac{1}{|\cC|}\sum_{c\in\cC}r_c^\top\mathrm{d}\theta_c$.
Let $a_k$ and $b_k$ be the gradients of $J$ with respect to $\theta_k$ and $v_k$.
In \cref{eq:supp_diff}, $\theta_k$ enters $\theta_{k+1}$, $v_{k+1}$ and, if $k\in\cC$, the objective, while $v_k$ enters $\theta_k$ and $v_{k+1}$.
Collecting these terms gives the backward recursion, for $k=K,\dots,1$,
\begin{equation}
\begin{aligned}
  a_k &= a_{k+1}+(H_k+\omega I)\,b_{k+1}+\frac{\mathbf{1}[k\in\cC]}{|\cC|}\,r_k,\\
  b_k &= \mu\,b_{k+1}-\eta_{k-1}\,a_k,
\end{aligned}
\label{eq:supp_adjoint}
\end{equation}
with $a_{K+1}=b_{K+1}=0$.
The text embeddings enter every step through $G_k$, so their gradient is a sum over the whole run:
\begin{equation}
  \nabla_T J=\sum_{k=0}^{K-1}G_k^\top\,b_{k+1}.
  \label{eq:supp_gradT}
\end{equation}
A truncated objective keeps only some of the terms of \cref{eq:supp_gradT}; TED keeps all $K$.
Each step of \cref{eq:supp_adjoint} needs one Hessian--vector product, so the backward pass costs a small multiple of the training run.

\paragraph{Gradient with respect to the displacement.}
TED keeps the norm of each initial embedding $t^0_m$ and optimizes its direction through a displacement $\delta_m$, initialized at zero:
\begin{equation}
  t_m=\|t^0_m\|\,\frac{u^0_m+\delta_m}{\|u^0_m+\delta_m\|},
  \qquad u^0_m=\frac{t^0_m}{\|t^0_m\|}.
  \label{eq:supp_param}
\end{equation}
Let $w_m=u^0_m+\delta_m$, so that $u_m=w_m/\|w_m\|$ and $t_m=\|t^0_m\|\,u_m$.
The Jacobian of the normalization is
\begin{equation}
  \frac{\partial u_m}{\partial\delta_m}=\frac{1}{\|w_m\|}\big(I-u_mu_m^\top\big),
  \label{eq:supp_jac}
\end{equation}
which is symmetric and removes the component along $u_m$.
The TED loss is $J+\lambda\Lanc$ with the anchor loss $\Lanc=\frac{1}{M}\sum_{m=1}^{M}\big(1-\langle u_m,u^0_m\rangle\big)$, and its two terms therefore have the gradients
\begin{equation}
\begin{aligned}
  \nabla_{\delta_m}J &= \frac{\|t^0_m\|}{\|w_m\|}\big(I-u_mu_m^\top\big)\nabla_{t_m}J,\\
  \nabla_{\delta_m}\Lanc &= -\frac{1}{M\|w_m\|}\big(I-u_mu_m^\top\big)u^0_m.
\end{aligned}
\label{eq:supp_graddelta}
\end{equation}
Both lie in the tangent space of the sphere at $u_m$: the first turns $u_m$ in the direction that lowers the real-data loss of the trained student, and the second turns it back towards $u^0_m$ and vanishes at $u_m=u^0_m$.
In practice, \cref{eq:supp_adjoint,eq:supp_gradT,eq:supp_graddelta} are computed by automatic differentiation over the full run, without truncation or implicit differentiation.
The memory grows linearly with $K$ and stays small, because the student is a projection head with about $1.2$M parameters.

\section{Implementation Details}
\label{sec:supp_impl}

\paragraph{Datasets.}
We use the Karpathy splits: Flickr30K with $29{,}000$ training, $1{,}014$ validation and $1{,}000$ test images, and MS-COCO with $113{,}287$, $5{,}000$ and $5{,}000$; every image has five captions.

\paragraph{Evaluation protocol.}
The student combines a frozen, pretrained image encoder with a trainable linear projection on the CLIP-L text embeddings.
It is trained with the symmetric InfoNCE loss at temperature $0.07$ by SGD with learning rate $0.2$, momentum $0.9$, weight decay $5\times10^{-4}$ and batch size $128$.
The number of epochs is set so that training takes about $100$ steps at every budget; the learning rate drops tenfold halfway, and the student is evaluated at about ten evenly spaced checkpoints.
We report R$_\text{mean}$ at the best checkpoint, averaged over five initializations.
All gains are differences of the reported values.
The TED surrogate uses the same projection, loss and schedule, with CLIP ViT-L/14 image features in place of the evaluation encoder.

\paragraph{TED solver.}
The surrogate student uses the same projection head, loss and schedule, with CLIP ViT-L/14 image features in place of the evaluation encoder, and $\cC$ is the set of evaluation checkpoints of the protocol.
The displacement $\delta$ is optimized with Adam for $300$ outer steps.
Every outer step samples four student initializations $\theta_0$ and a batch $\cB$ of $4{,}096$ real training images, each paired with one of its captions at random.
A held-out split of $5\%$ of the training images is excluded from these batches; every $25$ outer steps, the solver evaluates $J+\lambda\Lanc$ on this split with four fixed initializations and keeps the solution of the selected step, the one with the lowest value.
The anchor weight $\lambda{=}3$ is chosen from $\{0,1,3,10\}$ on the validation split and used for every dataset, budget and synthetic set.

\section{Synthetic Sets and Text Baselines}
\label{sec:supp_sets}

\paragraph{PDS.}
We run the public code of PDS~\cite{choi2026multi} on Flickr30K and MS-COCO for $M\in\{100,300,500,1000\}$. PDS clusters the CLIP-L image and text embeddings of the training set, matches image and text clusters, and decodes each image prototype with unCLIP. The upstream text embedding $t^0_m$ is the center of the matched text cluster.

\paragraph{EDGE.}
We generate the EDGE sets~\cite{zhao2025edge} with the settings of the original paper. The upstream text embedding $t^0_m$ is the CLIP-L embedding of the caption that EDGE pairs with each generated image.

\paragraph{Nearest real images.}
For each PDS image prototype, the real-image set of \cref{tab:realimg} takes the nearest real training image in CLIP-L image space, with the 100 images all distinct, and keeps the PDS text embedding of that prototype.

\paragraph{Training-free text embeddings.}
The heuristics of \cref{tab:supp_heuristics} replace $t^0_m$ by an embedding computed from the generated image $\tilde x_m$ without any student:
\textit{Best caption}, the real training caption with the highest CLIP-L similarity to $\tilde x_m$;
\textit{$k$-NN}, the mean caption embedding of the $k$ real training images nearest to $\tilde x_m$, for $k\in\{1,5,20\}$;
\textit{Image embedding}, the CLIP-L image embedding of $\tilde x_m$ itself;
and \textit{Shift}, $t^0_m$ moved halfway towards the 5-NN embedding.
All of them are rescaled to the norm of $t^0_m$.

\paragraph{Trajectory-matching objective.}
To isolate the role of the objective, this variant keeps the images, the surrogate, the parametrization of \cref{eq:supp_param} and the anchor loss of TED, and replaces $J$ by the trajectory-matching loss of MTT-VL and LoRS~\cite{wu2024multi,xu2024lors}: ten experts are trained for ten epochs on all real training pairs in the surrogate space, and the student is trained on the synthetic pairs from an expert checkpoint for $N$ steps and matched to the expert parameters $E$ epochs later. The configuration $(N,E,S_{\max})\in\{(8,1,2),(20,2,2),(50,2,4)\}$, where $S_{\max}$ is the latest starting epoch, is selected on the validation split separately for $\lambda{=}0$ and $\lambda{=}3$, and the solution of the last outer step is evaluated.

\section{Full Results}
\label{sec:supp_full}

\paragraph{Scaling the budget.}
\Cref{tab:supp_scaling} reports PDS and PDS + TED from $M{=}100$ to $M{=}1000$ and shows that TED improves PDS at every budget on both datasets and both architectures, with larger gains on Flickr30K than on MS-COCO.
The improvement is large enough that, on Flickr30K with ResNet, PDS + TED with 500 pairs already outperforms PDS with twice as many pairs, and on MS-COCO with ViT the gain even grows as the budget increases.
Generating more images widens the coverage of the training distribution but again assigns each new image a heuristic text, whereas TED improves what each pair teaches, so that optimizing the text complements generating more images rather than replacing it.

\begin{table}[!tb]
\centering
\footnotesize
\setlength{\tabcolsep}{2.5pt}
\caption{\textbf{Scaling the budget}. Brackets: gain over PDS.}
\label{tab:supp_scaling}
\setlength{\aboverulesep}{0pt}
\setlength{\belowrulesep}{0pt}
\setlength{\extrarowheight}{2pt}
\begin{tabular}{ll|cccc}
\toprule
Backbone & Method & $M{=}100$ & $300$ & $500$ & $1000$ \\
\midrule
\multicolumn{6}{l}{\textit{Flickr30K}} \\
\midrule
\multirow{2}{*}{ResNet} & PDS & 25.1 & 38.0 & 40.8 & 44.2 \\
 & \cellcolor[gray]{0.93}PDS + TED & \cellcolor[gray]{0.93}29.5 {\scriptsize(+4.4)} & \cellcolor[gray]{0.93}41.6 {\scriptsize(+3.6)} & \cellcolor[gray]{0.93}44.7 {\scriptsize(+3.9)} & \cellcolor[gray]{0.93}47.9 {\scriptsize(+3.7)} \\
\cmidrule(lr){1-6}
\multirow{2}{*}{ViT} & PDS & 15.1 & 24.5 & 28.6 & 32.0 \\
 & \cellcolor[gray]{0.93}PDS + TED & \cellcolor[gray]{0.93}19.3 {\scriptsize(+4.2)} & \cellcolor[gray]{0.93}27.6 {\scriptsize(+3.1)} & \cellcolor[gray]{0.93}31.7 {\scriptsize(+3.1)} & \cellcolor[gray]{0.93}35.1 {\scriptsize(+3.1)} \\
\midrule
\multicolumn{6}{l}{\textit{MS-COCO}} \\
\midrule
\multirow{2}{*}{ResNet} & PDS & 11.5 & 16.4 & 19.2 & 22.5 \\
 & \cellcolor[gray]{0.93}PDS + TED & \cellcolor[gray]{0.93}12.8 {\scriptsize(+1.3)} & \cellcolor[gray]{0.93}18.1 {\scriptsize(+1.7)} & \cellcolor[gray]{0.93}21.0 {\scriptsize(+1.8)} & \cellcolor[gray]{0.93}24.1 {\scriptsize(+1.6)} \\
\cmidrule(lr){1-6}
\multirow{2}{*}{ViT} & PDS & 7.8 & 11.7 & 13.9 & 15.6 \\
 & \cellcolor[gray]{0.93}PDS + TED & \cellcolor[gray]{0.93}9.1 {\scriptsize(+1.3)} & \cellcolor[gray]{0.93}13.5 {\scriptsize(+1.8)} & \cellcolor[gray]{0.93}16.1 {\scriptsize(+2.2)} & \cellcolor[gray]{0.93}18.0 {\scriptsize(+2.4)} \\
\bottomrule
\end{tabular}
\end{table}

\paragraph{Cross-architecture evaluation.}
\Cref{tab:supp_nfnet} evaluates PDS and PDS + TED with an NFNet student, the architecture on which TESLA-VL and LoRS are distilled and which TED never sees during optimization.
TED improves PDS in all settings, with larger gains on Flickr30K than on MS-COCO as for ResNet and ViT, and on MS-COCO the gain again grows with the budget, as it does for ViT in \cref{tab:supp_scaling}.
The text embeddings optimized with the CLIP surrogate therefore transfer to a third family of image encoders.

\begin{table}[!tb]
\centering
\scriptsize
\setlength{\tabcolsep}{2.2pt}
\setlength{\aboverulesep}{0pt}
\setlength{\belowrulesep}{0pt}
\setlength{\extrarowheight}{2pt}
\caption{\textbf{NFNet students}. R$_\text{mean}$ (\%), mean over five initializations. Brackets: gain over PDS.}
\label{tab:supp_nfnet}
\begin{tabularx}{\columnwidth}{ll|YYYY}
\toprule
Dataset & Method & $M{=}100$ & $M{=}300$ & $M{=}500$ & $M{=}1000$ \\
\midrule
\multirow{2}{*}{Flickr30K} & PDS & 26.4 & 41.6 & 46.4 & 50.8 \\
 & \cellcolor[gray]{0.93}PDS + TED & \cellcolor[gray]{0.93}31.0 {\scriptsize(+4.6)} & \cellcolor[gray]{0.93}45.1 {\scriptsize(+3.5)} & \cellcolor[gray]{0.93}49.9 {\scriptsize(+3.5)} & \cellcolor[gray]{0.93}54.8 {\scriptsize(+4.0)} \\
\midrule
\multirow{2}{*}{MS-COCO} & PDS & 9.9 & 14.9 & 18.0 & 22.3 \\
 & \cellcolor[gray]{0.93}PDS + TED & \cellcolor[gray]{0.93}10.6 {\scriptsize(+0.7)} & \cellcolor[gray]{0.93}16.2 {\scriptsize(+1.3)} & \cellcolor[gray]{0.93}19.5 {\scriptsize(+1.5)} & \cellcolor[gray]{0.93}24.3 {\scriptsize(+2.0)} \\
\bottomrule
\end{tabularx}
\end{table}

\paragraph{All retrieval metrics.}
\Cref{tab:supp_full_flickr,tab:supp_full_coco} report the six recalls and R$_\text{mean}$ with standard deviations over five student initializations for every generative method, budget and architecture, including $M{=}1000$ for PDS, and show that the variation across initializations is small compared with the gain of TED in every setting.
TED raises every recall in every setting, and the gain grows with the cutoff and is larger for text retrieval than for image retrieval, so that TED mainly brings more correct captions into the top of each ranking.

\begin{table*}[!t]
\centering
\scriptsize
\setlength{\aboverulesep}{0pt}
\setlength{\belowrulesep}{0pt}
\setlength{\extrarowheight}{2pt}
\caption{\textbf{Full results on Flickr30K}. Recall@$k$ (\%) and R$_\text{mean}$, mean $\pm$ std over five student initializations.}
\label{tab:supp_full_flickr}
\resizebox{\textwidth}{!}{%
\begin{tabular}{cl|ccc|ccc|c}
\toprule
$M$ & Method & IR@1 & IR@5 & IR@10 & TR@1 & TR@5 & TR@10 & R$_\text{mean}$ \\
\midrule
\multicolumn{9}{l}{\textit{ResNet}} \\
\midrule
\multirow{4}{*}{100} & EDGE & 6.0$\pm$0.2 & 19.6$\pm$0.2 & 29.1$\pm$0.3 & 7.6$\pm$0.6 & 21.3$\pm$0.9 & 30.9$\pm$0.4 & 19.1$\pm$0.3 \\
 & \cellcolor[gray]{0.93}EDGE + TED & \cellcolor[gray]{0.93}7.9$\pm$0.2 & \cellcolor[gray]{0.93}24.3$\pm$0.2 & \cellcolor[gray]{0.93}35.3$\pm$0.3 & \cellcolor[gray]{0.93}11.3$\pm$0.5 & \cellcolor[gray]{0.93}27.1$\pm$0.6 & \cellcolor[gray]{0.93}38.0$\pm$0.6 & \cellcolor[gray]{0.93}24.0$\pm$0.3 \\
 & PDS & 8.4$\pm$0.1 & 25.6$\pm$0.3 & 36.7$\pm$0.2 & 11.6$\pm$0.4 & 29.1$\pm$0.4 & 39.3$\pm$0.6 & 25.1$\pm$0.2 \\
 & \cellcolor[gray]{0.93}PDS + TED & \cellcolor[gray]{0.93}10.2$\pm$0.1 & \cellcolor[gray]{0.93}29.7$\pm$0.2 & \cellcolor[gray]{0.93}42.1$\pm$0.3 & \cellcolor[gray]{0.93}14.5$\pm$0.5 & \cellcolor[gray]{0.93}34.3$\pm$0.3 & \cellcolor[gray]{0.93}45.9$\pm$0.4 & \cellcolor[gray]{0.93}29.5$\pm$0.2 \\
\cmidrule(lr){1-9}
\multirow{4}{*}{300} & EDGE & 12.7$\pm$0.2 & 33.9$\pm$0.3 & 46.2$\pm$0.4 & 15.8$\pm$0.5 & 38.9$\pm$0.6 & 51.5$\pm$0.6 & 33.2$\pm$0.1 \\
 & \cellcolor[gray]{0.93}EDGE + TED & \cellcolor[gray]{0.93}14.3$\pm$0.2 & \cellcolor[gray]{0.93}36.6$\pm$0.3 & \cellcolor[gray]{0.93}49.1$\pm$0.2 & \cellcolor[gray]{0.93}19.4$\pm$0.9 & \cellcolor[gray]{0.93}43.2$\pm$0.6 & \cellcolor[gray]{0.93}56.4$\pm$0.7 & \cellcolor[gray]{0.93}36.5$\pm$0.3 \\
 & PDS & 14.8$\pm$0.2 & 38.3$\pm$0.2 & 50.4$\pm$0.2 & 22.6$\pm$0.6 & 44.9$\pm$0.5 & 56.9$\pm$0.6 & 38.0$\pm$0.2 \\
 & \cellcolor[gray]{0.93}PDS + TED & \cellcolor[gray]{0.93}16.4$\pm$0.2 & \cellcolor[gray]{0.93}41.8$\pm$0.2 & \cellcolor[gray]{0.93}54.9$\pm$0.1 & \cellcolor[gray]{0.93}24.3$\pm$1.0 & \cellcolor[gray]{0.93}49.5$\pm$0.3 & \cellcolor[gray]{0.93}62.8$\pm$0.5 & \cellcolor[gray]{0.93}41.6$\pm$0.2 \\
\cmidrule(lr){1-9}
\multirow{4}{*}{500} & EDGE & 14.7$\pm$0.3 & 37.3$\pm$0.2 & 50.6$\pm$0.2 & 20.1$\pm$0.6 & 44.2$\pm$0.9 & 56.4$\pm$0.8 & 37.2$\pm$0.2 \\
 & \cellcolor[gray]{0.93}EDGE + TED & \cellcolor[gray]{0.93}16.8$\pm$0.3 & \cellcolor[gray]{0.93}40.7$\pm$0.2 & \cellcolor[gray]{0.93}53.4$\pm$0.3 & \cellcolor[gray]{0.93}24.2$\pm$0.4 & \cellcolor[gray]{0.93}49.6$\pm$0.7 & \cellcolor[gray]{0.93}62.3$\pm$0.7 & \cellcolor[gray]{0.93}41.1$\pm$0.2 \\
 & PDS & 16.8$\pm$0.2 & 40.5$\pm$0.2 & 54.6$\pm$0.3 & 23.2$\pm$0.5 & 49.0$\pm$0.3 & 60.5$\pm$0.5 & 40.8$\pm$0.1 \\
 & \cellcolor[gray]{0.93}PDS + TED & \cellcolor[gray]{0.93}18.2$\pm$0.2 & \cellcolor[gray]{0.93}43.7$\pm$0.2 & \cellcolor[gray]{0.93}57.5$\pm$0.2 & \cellcolor[gray]{0.93}26.2$\pm$0.6 & \cellcolor[gray]{0.93}55.4$\pm$0.4 & \cellcolor[gray]{0.93}66.8$\pm$0.9 & \cellcolor[gray]{0.93}44.7$\pm$0.2 \\
\cmidrule(lr){1-9}
\multirow{2}{*}{1000} & PDS & 17.6$\pm$0.2 & 43.8$\pm$0.3 & 57.1$\pm$0.2 & 26.2$\pm$0.9 & 53.9$\pm$0.5 & 66.4$\pm$0.8 & 44.2$\pm$0.2 \\
 & \cellcolor[gray]{0.93}PDS + TED & \cellcolor[gray]{0.93}19.4$\pm$0.3 & \cellcolor[gray]{0.93}47.1$\pm$0.2 & \cellcolor[gray]{0.93}60.7$\pm$0.1 & \cellcolor[gray]{0.93}30.5$\pm$0.5 & \cellcolor[gray]{0.93}58.7$\pm$0.6 & \cellcolor[gray]{0.93}71.0$\pm$0.6 & \cellcolor[gray]{0.93}47.9$\pm$0.3 \\
\midrule
\multicolumn{9}{l}{\textit{ViT}} \\
\midrule
\multirow{4}{*}{100} & EDGE & 4.2$\pm$0.2 & 13.4$\pm$0.3 & 20.7$\pm$0.4 & 5.1$\pm$0.4 & 13.4$\pm$0.6 & 20.7$\pm$0.8 & 12.9$\pm$0.2 \\
 & \cellcolor[gray]{0.93}EDGE + TED & \cellcolor[gray]{0.93}4.9$\pm$0.1 & \cellcolor[gray]{0.93}16.8$\pm$0.3 & \cellcolor[gray]{0.93}25.8$\pm$0.3 & \cellcolor[gray]{0.93}6.4$\pm$0.2 & \cellcolor[gray]{0.93}17.4$\pm$0.5 & \cellcolor[gray]{0.93}24.4$\pm$0.4 & \cellcolor[gray]{0.93}16.0$\pm$0.2 \\
 & PDS & 6.1$\pm$0.2 & 17.5$\pm$0.4 & 26.0$\pm$0.3 & 4.6$\pm$0.2 & 14.7$\pm$0.7 & 22.0$\pm$0.3 & 15.1$\pm$0.1 \\
 & \cellcolor[gray]{0.93}PDS + TED & \cellcolor[gray]{0.93}7.4$\pm$0.2 & \cellcolor[gray]{0.93}21.0$\pm$0.3 & \cellcolor[gray]{0.93}31.6$\pm$0.3 & \cellcolor[gray]{0.93}7.5$\pm$0.4 & \cellcolor[gray]{0.93}20.0$\pm$0.2 & \cellcolor[gray]{0.93}28.4$\pm$0.4 & \cellcolor[gray]{0.93}19.3$\pm$0.1 \\
\cmidrule(lr){1-9}
\multirow{4}{*}{300} & EDGE & 7.5$\pm$0.2 & 22.8$\pm$0.3 & 33.0$\pm$0.4 & 9.1$\pm$0.6 & 24.5$\pm$0.5 & 34.0$\pm$0.8 & 21.8$\pm$0.2 \\
 & \cellcolor[gray]{0.93}EDGE + TED & \cellcolor[gray]{0.93}9.2$\pm$0.1 & \cellcolor[gray]{0.93}26.5$\pm$0.2 & \cellcolor[gray]{0.93}37.4$\pm$0.3 & \cellcolor[gray]{0.93}9.8$\pm$0.8 & \cellcolor[gray]{0.93}26.5$\pm$0.5 & \cellcolor[gray]{0.93}37.9$\pm$1.0 & \cellcolor[gray]{0.93}24.5$\pm$0.2 \\
 & PDS & 9.6$\pm$0.3 & 25.9$\pm$0.5 & 36.9$\pm$0.3 & 10.3$\pm$0.5 & 27.0$\pm$0.7 & 37.5$\pm$0.8 & 24.5$\pm$0.1 \\
 & \cellcolor[gray]{0.93}PDS + TED & \cellcolor[gray]{0.93}11.1$\pm$0.2 & \cellcolor[gray]{0.93}29.3$\pm$0.4 & \cellcolor[gray]{0.93}40.6$\pm$0.6 & \cellcolor[gray]{0.93}12.8$\pm$0.9 & \cellcolor[gray]{0.93}30.6$\pm$0.5 & \cellcolor[gray]{0.93}41.4$\pm$0.8 & \cellcolor[gray]{0.93}27.6$\pm$0.2 \\
\cmidrule(lr){1-9}
\multirow{4}{*}{500} & EDGE & 9.6$\pm$0.2 & 27.1$\pm$0.3 & 37.9$\pm$0.5 & 11.2$\pm$0.2 & 28.9$\pm$0.3 & 40.1$\pm$0.6 & 25.8$\pm$0.1 \\
 & \cellcolor[gray]{0.93}EDGE + TED & \cellcolor[gray]{0.93}11.3$\pm$0.1 & \cellcolor[gray]{0.93}31.0$\pm$0.4 & \cellcolor[gray]{0.93}42.3$\pm$0.3 & \cellcolor[gray]{0.93}12.7$\pm$0.6 & \cellcolor[gray]{0.93}32.3$\pm$0.8 & \cellcolor[gray]{0.93}43.9$\pm$0.5 & \cellcolor[gray]{0.93}28.9$\pm$0.2 \\
 & PDS & 11.5$\pm$0.3 & 30.0$\pm$0.3 & 42.0$\pm$0.3 & 12.5$\pm$0.9 & 32.1$\pm$0.7 & 43.4$\pm$0.4 & 28.6$\pm$0.2 \\
 & \cellcolor[gray]{0.93}PDS + TED & \cellcolor[gray]{0.93}12.6$\pm$0.2 & \cellcolor[gray]{0.93}32.8$\pm$0.5 & \cellcolor[gray]{0.93}44.6$\pm$0.3 & \cellcolor[gray]{0.93}15.0$\pm$0.4 & \cellcolor[gray]{0.93}36.7$\pm$0.5 & \cellcolor[gray]{0.93}48.2$\pm$0.7 & \cellcolor[gray]{0.93}31.7$\pm$0.2 \\
\cmidrule(lr){1-9}
\multirow{2}{*}{1000} & PDS & 13.4$\pm$0.4 & 33.3$\pm$0.3 & 44.6$\pm$0.2 & 15.9$\pm$0.5 & 36.4$\pm$0.3 & 48.5$\pm$0.6 & 32.0$\pm$0.1 \\
 & \cellcolor[gray]{0.93}PDS + TED & \cellcolor[gray]{0.93}14.8$\pm$0.4 & \cellcolor[gray]{0.93}35.7$\pm$0.0 & \cellcolor[gray]{0.93}47.1$\pm$0.3 & \cellcolor[gray]{0.93}19.5$\pm$0.5 & \cellcolor[gray]{0.93}40.9$\pm$1.2 & \cellcolor[gray]{0.93}52.6$\pm$0.6 & \cellcolor[gray]{0.93}35.1$\pm$0.4 \\
\bottomrule
\end{tabular}%
}
\end{table*}

\begin{table*}[!t]
\centering
\scriptsize
\caption{\textbf{Full results on MS-COCO}. Recall@$k$ (\%) and R$_\text{mean}$, mean $\pm$ std over five student initializations.}
\label{tab:supp_full_coco}
\resizebox{\textwidth}{!}{%
\begin{tabular}{cl|ccc|ccc|c}
\toprule
$M$ & Method & IR@1 & IR@5 & IR@10 & TR@1 & TR@5 & TR@10 & R$_\text{mean}$ \\
\midrule
\multicolumn{9}{l}{\textit{ResNet}} \\
\midrule
\multirow{4}{*}{100} & EDGE & 2.1$\pm$0.0 & 7.4$\pm$0.1 & 12.4$\pm$0.1 & 2.7$\pm$0.2 & 8.8$\pm$0.3 & 14.5$\pm$0.3 & 8.0$\pm$0.1 \\
 & \cellcolor[gray]{0.93}EDGE + TED & \cellcolor[gray]{0.93}2.4$\pm$0.0 & \cellcolor[gray]{0.93}8.3$\pm$0.0 & \cellcolor[gray]{0.93}14.2$\pm$0.1 & \cellcolor[gray]{0.93}3.7$\pm$0.2 & \cellcolor[gray]{0.93}12.6$\pm$0.2 & \cellcolor[gray]{0.93}19.7$\pm$0.2 & \cellcolor[gray]{0.93}10.2$\pm$0.1 \\
 & PDS & 2.9$\pm$0.0 & 10.5$\pm$0.1 & 17.7$\pm$0.1 & 3.9$\pm$0.2 & 13.3$\pm$0.3 & 21.0$\pm$0.3 & 11.5$\pm$0.1 \\
 & \cellcolor[gray]{0.93}PDS + TED & \cellcolor[gray]{0.93}3.1$\pm$0.1 & \cellcolor[gray]{0.93}11.3$\pm$0.1 & \cellcolor[gray]{0.93}18.7$\pm$0.1 & \cellcolor[gray]{0.93}4.9$\pm$0.2 & \cellcolor[gray]{0.93}15.0$\pm$0.2 & \cellcolor[gray]{0.93}23.6$\pm$0.2 & \cellcolor[gray]{0.93}12.8$\pm$0.0 \\
\cmidrule(lr){1-9}
\multirow{4}{*}{300} & EDGE & 3.9$\pm$0.1 & 12.8$\pm$0.1 & 20.6$\pm$0.2 & 5.8$\pm$0.2 & 17.4$\pm$0.1 & 26.1$\pm$0.3 & 14.4$\pm$0.1 \\
 & \cellcolor[gray]{0.93}EDGE + TED & \cellcolor[gray]{0.93}4.5$\pm$0.1 & \cellcolor[gray]{0.93}14.4$\pm$0.1 & \cellcolor[gray]{0.93}22.8$\pm$0.2 & \cellcolor[gray]{0.93}7.6$\pm$0.2 & \cellcolor[gray]{0.93}22.2$\pm$0.3 & \cellcolor[gray]{0.93}31.8$\pm$0.4 & \cellcolor[gray]{0.93}17.2$\pm$0.1 \\
 & PDS & 4.6$\pm$0.1 & 15.1$\pm$0.1 & 23.9$\pm$0.1 & 6.6$\pm$0.1 & 19.2$\pm$0.4 & 28.7$\pm$0.3 & 16.4$\pm$0.1 \\
 & \cellcolor[gray]{0.93}PDS + TED & \cellcolor[gray]{0.93}5.0$\pm$0.1 & \cellcolor[gray]{0.93}15.9$\pm$0.1 & \cellcolor[gray]{0.93}25.3$\pm$0.1 & \cellcolor[gray]{0.93}7.7$\pm$0.2 & \cellcolor[gray]{0.93}22.1$\pm$0.3 & \cellcolor[gray]{0.93}32.5$\pm$0.3 & \cellcolor[gray]{0.93}18.1$\pm$0.1 \\
\cmidrule(lr){1-9}
\multirow{4}{*}{500} & EDGE & 4.9$\pm$0.1 & 15.6$\pm$0.2 & 24.5$\pm$0.3 & 7.6$\pm$0.3 & 22.1$\pm$0.2 & 32.1$\pm$0.2 & 17.8$\pm$0.1 \\
 & \cellcolor[gray]{0.93}EDGE + TED & \cellcolor[gray]{0.93}5.3$\pm$0.1 & \cellcolor[gray]{0.93}16.7$\pm$0.2 & \cellcolor[gray]{0.93}26.3$\pm$0.1 & \cellcolor[gray]{0.93}9.4$\pm$0.2 & \cellcolor[gray]{0.93}25.3$\pm$0.2 & \cellcolor[gray]{0.93}36.1$\pm$0.4 & \cellcolor[gray]{0.93}19.9$\pm$0.1 \\
 & PDS & 5.1$\pm$0.0 & 17.0$\pm$0.2 & 26.8$\pm$0.3 & 8.4$\pm$0.1 & 23.8$\pm$0.3 & 34.2$\pm$0.5 & 19.2$\pm$0.1 \\
 & \cellcolor[gray]{0.93}PDS + TED & \cellcolor[gray]{0.93}5.7$\pm$0.1 & \cellcolor[gray]{0.93}18.2$\pm$0.1 & \cellcolor[gray]{0.93}28.4$\pm$0.1 & \cellcolor[gray]{0.93}9.9$\pm$0.2 & \cellcolor[gray]{0.93}26.5$\pm$0.2 & \cellcolor[gray]{0.93}37.6$\pm$0.4 & \cellcolor[gray]{0.93}21.0$\pm$0.1 \\
\cmidrule(lr){1-9}
\multirow{2}{*}{1000} & PDS & 6.4$\pm$0.0 & 19.8$\pm$0.2 & 30.1$\pm$0.1 & 10.8$\pm$0.2 & 27.9$\pm$0.3 & 39.7$\pm$0.4 & 22.5$\pm$0.1 \\
 & \cellcolor[gray]{0.93}PDS + TED & \cellcolor[gray]{0.93}6.7$\pm$0.1 & \cellcolor[gray]{0.93}20.4$\pm$0.1 & \cellcolor[gray]{0.93}31.2$\pm$0.1 & \cellcolor[gray]{0.93}12.3$\pm$0.2 & \cellcolor[gray]{0.93}31.1$\pm$0.3 & \cellcolor[gray]{0.93}43.0$\pm$0.3 & \cellcolor[gray]{0.93}24.1$\pm$0.1 \\
\midrule
\multicolumn{9}{l}{\textit{ViT}} \\
\midrule
\multirow{4}{*}{100} & EDGE & 1.2$\pm$0.1 & 4.7$\pm$0.3 & 8.4$\pm$0.4 & 1.1$\pm$0.1 & 4.6$\pm$0.3 & 7.9$\pm$0.4 & 4.6$\pm$0.1 \\
 & \cellcolor[gray]{0.93}EDGE + TED & \cellcolor[gray]{0.93}1.9$\pm$0.0 & \cellcolor[gray]{0.93}6.8$\pm$0.1 & \cellcolor[gray]{0.93}11.6$\pm$0.2 & \cellcolor[gray]{0.93}1.8$\pm$0.1 & \cellcolor[gray]{0.93}6.2$\pm$0.3 & \cellcolor[gray]{0.93}10.8$\pm$0.2 & \cellcolor[gray]{0.93}6.5$\pm$0.1 \\
 & PDS & 2.2$\pm$0.0 & 8.1$\pm$0.1 & 13.8$\pm$0.1 & 2.4$\pm$0.1 & 7.6$\pm$0.1 & 12.7$\pm$0.3 & 7.8$\pm$0.1 \\
 & \cellcolor[gray]{0.93}PDS + TED & \cellcolor[gray]{0.93}2.5$\pm$0.1 & \cellcolor[gray]{0.93}9.1$\pm$0.1 & \cellcolor[gray]{0.93}15.5$\pm$0.2 & \cellcolor[gray]{0.93}2.6$\pm$0.1 & \cellcolor[gray]{0.93}9.6$\pm$0.3 & \cellcolor[gray]{0.93}15.4$\pm$0.4 & \cellcolor[gray]{0.93}9.1$\pm$0.1 \\
\cmidrule(lr){1-9}
\multirow{4}{*}{300} & EDGE & 3.3$\pm$0.1 & 10.7$\pm$0.1 & 17.2$\pm$0.1 & 2.8$\pm$0.1 & 10.0$\pm$0.1 & 16.0$\pm$0.3 & 10.0$\pm$0.1 \\
 & \cellcolor[gray]{0.93}EDGE + TED & \cellcolor[gray]{0.93}3.5$\pm$0.0 & \cellcolor[gray]{0.93}12.1$\pm$0.0 & \cellcolor[gray]{0.93}19.5$\pm$0.1 & \cellcolor[gray]{0.93}3.7$\pm$0.1 & \cellcolor[gray]{0.93}12.6$\pm$0.3 & \cellcolor[gray]{0.93}19.7$\pm$0.4 & \cellcolor[gray]{0.93}11.8$\pm$0.1 \\
 & PDS & 3.7$\pm$0.1 & 12.5$\pm$0.1 & 20.3$\pm$0.1 & 3.4$\pm$0.1 & 11.7$\pm$0.3 & 18.7$\pm$0.3 & 11.7$\pm$0.1 \\
 & \cellcolor[gray]{0.93}PDS + TED & \cellcolor[gray]{0.93}4.1$\pm$0.0 & \cellcolor[gray]{0.93}14.0$\pm$0.2 & \cellcolor[gray]{0.93}22.2$\pm$0.1 & \cellcolor[gray]{0.93}4.5$\pm$0.1 & \cellcolor[gray]{0.93}14.4$\pm$0.4 & \cellcolor[gray]{0.93}22.1$\pm$0.4 & \cellcolor[gray]{0.93}13.5$\pm$0.2 \\
\cmidrule(lr){1-9}
\multirow{4}{*}{500} & EDGE & 3.9$\pm$0.1 & 12.7$\pm$0.1 & 20.2$\pm$0.2 & 3.5$\pm$0.2 & 12.0$\pm$0.4 & 19.3$\pm$0.6 & 11.9$\pm$0.2 \\
 & \cellcolor[gray]{0.93}EDGE + TED & \cellcolor[gray]{0.93}4.4$\pm$0.0 & \cellcolor[gray]{0.93}14.3$\pm$0.1 & \cellcolor[gray]{0.93}22.3$\pm$0.1 & \cellcolor[gray]{0.93}4.7$\pm$0.1 & \cellcolor[gray]{0.93}14.1$\pm$0.4 & \cellcolor[gray]{0.93}21.9$\pm$0.4 & \cellcolor[gray]{0.93}13.6$\pm$0.2 \\
 & PDS & 4.5$\pm$0.1 & 14.4$\pm$0.1 & 22.5$\pm$0.1 & 4.3$\pm$0.1 & 14.7$\pm$0.1 & 22.9$\pm$0.3 & 13.9$\pm$0.1 \\
 & \cellcolor[gray]{0.93}PDS + TED & \cellcolor[gray]{0.93}5.0$\pm$0.1 & \cellcolor[gray]{0.93}16.2$\pm$0.1 & \cellcolor[gray]{0.93}24.7$\pm$0.2 & \cellcolor[gray]{0.93}5.9$\pm$0.1 & \cellcolor[gray]{0.93}17.9$\pm$0.2 & \cellcolor[gray]{0.93}26.8$\pm$0.3 & \cellcolor[gray]{0.93}16.1$\pm$0.1 \\
\cmidrule(lr){1-9}
\multirow{2}{*}{1000} & PDS & 5.0$\pm$0.1 & 16.1$\pm$0.2 & 24.6$\pm$0.1 & 5.7$\pm$0.1 & 16.6$\pm$0.2 & 25.4$\pm$0.6 & 15.6$\pm$0.1 \\
 & \cellcolor[gray]{0.93}PDS + TED & \cellcolor[gray]{0.93}5.8$\pm$0.1 & \cellcolor[gray]{0.93}17.9$\pm$0.1 & \cellcolor[gray]{0.93}27.0$\pm$0.2 & \cellcolor[gray]{0.93}7.1$\pm$0.2 & \cellcolor[gray]{0.93}20.4$\pm$0.2 & \cellcolor[gray]{0.93}30.0$\pm$0.4 & \cellcolor[gray]{0.93}18.0$\pm$0.1 \\
\bottomrule
\end{tabular}%
}
\end{table*}

\paragraph{Variance over solves.}
\Cref{tab:supp_seeds} repeats every PDS solve with two further solver seeds, which change the student initializations, the real batches and the held-out split used by the solver; the images and $t^0$ stay the same.
Across the three solves, R$_\text{mean}$ has a standard deviation of $0.01$ to $0.12$ and a range of at most $0.22$, one to two orders of magnitude below the gains of $+1.22$ to $+4.40$.
Every solve improves every setting, and the least favorable solve still gains $+1.22$.
The mean cosine between $t^\star$ and $t^0$ agrees across seeds to within $0.001$ in every setting: $\Lanc$ fixes how far the text moves, and where exactly it ends within that range has no measurable effect on the student.
A single solve is therefore representative, and all other results use seed 0.

\begin{table*}[!t]
\centering
\footnotesize
\setlength{\aboverulesep}{0pt}
\setlength{\belowrulesep}{0pt}
\setlength{\extrarowheight}{2pt}
\caption{\textbf{TED over three solver seeds} (PDS, $\lambda{=}3$). R$_\text{mean}$ (\%) of each solve, mean $\pm$ std over five student initializations. Seed 0 is the solve used in all other results.}
\label{tab:supp_seeds}
\setlength{\tabcolsep}{2.5pt}
\begin{tabularx}{\textwidth}{lc|YYYY|YYYY}
\toprule
 & & \multicolumn{4}{c|}{ResNet} & \multicolumn{4}{c}{ViT} \\
Dataset & $M$ & PDS & Seed 0 & Seed 1 & Seed 2 & PDS & Seed 0 & Seed 1 & Seed 2 \\
\midrule
Flickr30K & 100 & 25.1$\pm$0.2 & 29.5$\pm$0.2 & 29.5$\pm$0.2 & 29.5$\pm$0.2 & 15.1$\pm$0.1 & 19.3$\pm$0.1 & 19.4$\pm$0.2 & 19.4$\pm$0.1 \\
Flickr30K & 300 & 38.0$\pm$0.2 & 41.6$\pm$0.2 & 41.5$\pm$0.2 & 41.6$\pm$0.3 & 24.5$\pm$0.1 & 27.6$\pm$0.2 & 27.6$\pm$0.3 & 27.5$\pm$0.3 \\
Flickr30K & 500 & 40.8$\pm$0.1 & 44.7$\pm$0.2 & 44.6$\pm$0.3 & 44.8$\pm$0.2 & 28.6$\pm$0.2 & 31.7$\pm$0.2 & 31.6$\pm$0.2 & 31.8$\pm$0.2 \\
\midrule
MS-COCO & 100 & 11.5$\pm$0.1 & 12.8$\pm$0.0 & 12.8$\pm$0.0 & 12.8$\pm$0.0 & 7.8$\pm$0.1 & 9.1$\pm$0.1 & 9.1$\pm$0.1 & 9.1$\pm$0.1 \\
MS-COCO & 300 & 16.4$\pm$0.1 & 18.1$\pm$0.1 & 18.1$\pm$0.1 & 18.1$\pm$0.1 & 11.7$\pm$0.1 & 13.5$\pm$0.2 & 13.6$\pm$0.2 & 13.6$\pm$0.2 \\
MS-COCO & 500 & 19.2$\pm$0.1 & 21.0$\pm$0.1 & 21.0$\pm$0.1 & 21.0$\pm$0.1 & 13.9$\pm$0.1 & 16.1$\pm$0.1 & 16.0$\pm$0.1 & 16.1$\pm$0.1 \\
\bottomrule
\end{tabularx}
\end{table*}

\section{Beyond the PDS Protocol}
\label{sec:supp_beyond}

\paragraph{Protocol.}
The student of MTT-VL and LoRS~\cite{wu2024multi,xu2024lors} trains the image encoder, initialized from ImageNet, together with the linear text projection.
We follow its training loop for data without learned learning rates: SGD with learning rate $0.1$ for both parts, momentum $0.9$, weight decay $5\times10^{-4}$, batch size $128$ and $100$ epochs with a tenfold decay halfway, the symmetric InfoNCE loss, and the best of $11$ evaluation checkpoints, averaged over five initializations.
At $M{=}100$ this is $101$ steps, as in the PDS protocol; the differences are the trainable image encoder and the learning rate.
Besides ResNet and ViT, we evaluate NFNet, the architecture whose trajectories MTT-VL and LoRS match.

\begin{table*}[!t]
\centering
\footnotesize
\setlength{\aboverulesep}{0pt}
\setlength{\belowrulesep}{0pt}
\setlength{\extrarowheight}{2pt}
\caption{\textbf{Trained image encoder}. R$_\text{mean}$ (\%), mean $\pm$ std over five student initializations, when the image encoder is trained together with the text projection. PDS + TED uses the TED text solved with the frozen surrogate, without re-solving; $\Delta$: gain.}
\label{tab:supp_protocol_a}
\setlength{\tabcolsep}{3pt}
\begin{tabularx}{\textwidth}{lc|YYc|YYc|YYc}
\toprule
 & & \multicolumn{3}{c|}{NFNet} & \multicolumn{3}{c|}{ResNet} & \multicolumn{3}{c}{ViT} \\
Dataset & $M$ & PDS & PDS + TED & $\Delta$ & PDS & PDS + TED & $\Delta$ & PDS & PDS + TED & $\Delta$ \\
\midrule
Flickr30K & 100 & 25.6$\pm$0.3 & 29.3$\pm$0.1 & $+3.7$ & 24.6$\pm$0.2 & 28.6$\pm$0.3 & $+4.0$ & 15.0$\pm$0.1 & 19.0$\pm$0.2 & $+4.0$ \\
 & 300 & 41.2$\pm$0.4 & 44.7$\pm$0.1 & $+3.5$ & 37.4$\pm$0.1 & 41.0$\pm$0.2 & $+3.6$ & 24.4$\pm$0.4 & 27.2$\pm$0.2 & $+2.8$ \\
 & 500 & 46.4$\pm$0.2 & 49.8$\pm$0.2 & $+3.4$ & 40.5$\pm$0.1 & 44.3$\pm$0.2 & $+3.8$ & 27.8$\pm$0.4 & 30.9$\pm$0.2 & $+3.1$ \\
\midrule
MS-COCO & 100 & 9.7$\pm$0.1 & 10.1$\pm$0.1 & $+0.4$ & 11.2$\pm$0.1 & 12.3$\pm$0.1 & $+1.1$ & 7.8$\pm$0.0 & 8.8$\pm$0.1 & $+1.0$ \\
 & 300 & 14.9$\pm$0.1 & 16.1$\pm$0.1 & $+1.2$ & 16.0$\pm$0.1 & 17.7$\pm$0.1 & $+1.7$ & 11.5$\pm$0.1 & 13.4$\pm$0.1 & $+1.9$ \\
 & 500 & 18.8$\pm$0.1 & 20.0$\pm$0.2 & $+1.2$ & 19.3$\pm$0.0 & 21.0$\pm$0.1 & $+1.7$ & 13.9$\pm$0.0 & 15.8$\pm$0.1 & $+1.9$ \\
\bottomrule
\end{tabularx}
\end{table*}

\paragraph{Trained image encoder.}
\Cref{tab:supp_protocol_a} evaluates the PDS sets with CLIP-L text under this protocol, in which the image encoder is trained together with the text projection, and shows that TED improves PDS in every setting even though its text embeddings are optimized for a frozen surrogate.
On ResNet and ViT, the gains are of similar size to those with a frozen image encoder, while the smallest gains are observed for NFNet on MS-COCO.

\begin{table}[!tb]
\centering
\scriptsize
\setlength{\aboverulesep}{0pt}
\setlength{\belowrulesep}{0pt}
\setlength{\extrarowheight}{2pt}
\caption{\textbf{BERT text space}. R$_\text{mean}$ (\%), mean $\pm$ std over five student initializations. Same: the training of \cref{tab:supp_protocol_a}; Native: the learned learning rates, similarity matrix and loss of LoRS. LoRS has 99 and 499 pairs at nominal budgets $M{=}100$ and $500$.}
\label{tab:supp_bert}
\setlength{\tabcolsep}{2pt}
\begin{tabularx}{\columnwidth}{cll|YYY}
\toprule
$M$ & Synthetic set & Training & NFNet & ResNet & ViT \\
\midrule
\multirow{6}{*}{100} & PDS & Same & 11.2$\pm$0.2 & 9.2$\pm$0.2 & 2.6$\pm$1.2 \\
 & \cellcolor[gray]{0.93}PDS + TED & \cellcolor[gray]{0.93}Same & \cellcolor[gray]{0.93}19.0$\pm$0.1 & \cellcolor[gray]{0.93}18.1$\pm$0.3 & \cellcolor[gray]{0.93}12.4$\pm$0.3 \\
 & LoRS & Same & 23.3$\pm$0.3 & 7.4$\pm$0.1 & 4.6$\pm$0.0 \\
 & \cellcolor[gray]{0.93}LoRS + TED & \cellcolor[gray]{0.93}Same & \cellcolor[gray]{0.93}24.0$\pm$0.2 & \cellcolor[gray]{0.93}13.3$\pm$0.2 & \cellcolor[gray]{0.93}8.5$\pm$0.1 \\
 & LoRS & Native & 23.5$\pm$0.5 & 3.6$\pm$1.4 & 0.7$\pm$0.1 \\
 & \cellcolor[gray]{0.93}LoRS + TED & \cellcolor[gray]{0.93}Native & \cellcolor[gray]{0.93}24.3$\pm$0.5 & \cellcolor[gray]{0.93}6.0$\pm$0.4 & \cellcolor[gray]{0.93}0.8$\pm$0.2 \\
\midrule
\multirow{6}{*}{500} & PDS & Same & 24.4$\pm$0.2 & 21.2$\pm$0.1 & 13.1$\pm$1.0 \\
 & \cellcolor[gray]{0.93}PDS + TED & \cellcolor[gray]{0.93}Same & \cellcolor[gray]{0.93}26.6$\pm$0.2 & \cellcolor[gray]{0.93}23.1$\pm$0.4 & \cellcolor[gray]{0.93}15.8$\pm$0.4 \\
 & LoRS & Same & 26.2$\pm$0.5 & 13.3$\pm$0.2 & 6.8$\pm$0.2 \\
 & \cellcolor[gray]{0.93}LoRS + TED & \cellcolor[gray]{0.93}Same & \cellcolor[gray]{0.93}28.2$\pm$0.2 & \cellcolor[gray]{0.93}20.4$\pm$0.2 & \cellcolor[gray]{0.93}12.0$\pm$0.4 \\
 & LoRS & Native & 34.0$\pm$0.3 & 1.9$\pm$0.3 & 0.6$\pm$0.0 \\
 & \cellcolor[gray]{0.93}LoRS + TED & \cellcolor[gray]{0.93}Native & \cellcolor[gray]{0.93}30.0$\pm$0.5 & \cellcolor[gray]{0.93}2.3$\pm$0.4 & \cellcolor[gray]{0.93}0.6$\pm$0.1 \\
\bottomrule
\end{tabularx}
\end{table}

\paragraph{BERT text space.}
We use the BERT features of the LoRS release for all training and test captions, and obtain the PDS text $t^0$ in this space by rerunning the clustering of PDS, which reproduces its CLIP-L text embeddings exactly, and averaging the BERT features of the same captions.
TED is applied with the same settings and $\lambda{=}3$ without re-tuning for BERT, with a surrogate student that maps BERT text to CLIP-L image features and real pairs that consist of CLIP-L image features and BERT captions.
LoRS is reproduced with its official code and hyperparameters, which yield 99 and 499 pairs at the nominal budgets $M{=}100$ and $500$.
Each LoRS set is evaluated both with the training of the other rows (Same) and with its own learned learning rates, similarity matrix and weighted BCE loss (Native).
For LoRS + TED, TED keeps the LoRS images, starts from the learned LoRS text and uses the CLIP-L features of the LoRS images in the surrogate, and under Native the TED text replaces the LoRS text while everything else is kept.

\Cref{tab:supp_bert} reports the results on all three architectures.
Under the same training, TED improves LoRS on every architecture at both budgets, with much larger gains on ResNet and ViT than on NFNet, and although LoRS + TED surpasses PDS + TED on NFNet at both budgets, PDS + TED remains the stronger set on ResNet and ViT.
Under the native protocol, the results differ, since TED slightly improves LoRS on NFNet at the smaller budget but lowers it at the larger one, and native LoRS stays near chance on ViT with or without TED.
We attribute this to the native loss and similarity matrix, which are used in evaluation but are not modeled by the InfoNCE surrogate of TED, so the gains under the same training do not carry over uniformly to the native protocol.

\section{Additional Ablations}
\label{sec:supp_ablation}

\paragraph{All retrieval metrics.}
\Cref{tab:supp_abl_full} reports the six recalls of the ablation variants on Flickr30K with $M{=}300$, together with the remaining training-free text embeddings, and shows that TED is the best variant on almost every metric, the only exception being IR@1 on ViT, where the variant without $\Lanc$ is marginally higher.
On ResNet, the gain of TED is considerably larger at the higher cutoffs than at IR@1 and TR@1.
Without $\Lanc$, the ResNet student even falls below PDS at IR@1 and TR@1 while improving at IR@10, which indicates that the unconstrained text changes the ranking without improving it on an image encoder other than CLIP, whereas adding $\Lanc$ to the same unrolled objective yields gains at every cutoff.

\begin{table*}[!t]
\centering
\footnotesize
\setlength{\aboverulesep}{0pt}
\setlength{\belowrulesep}{0pt}
\setlength{\extrarowheight}{2pt}
\caption{\textbf{Ablations with all retrieval metrics}. Recall@$k$ (\%) and R$_\text{mean}$, mean $\pm$ std over five student initializations. One inner step: TED with $K{=}1$; full unroll w/o $\Lanc$: TED with $\lambda{=}0$.}
\label{tab:supp_abl_full}
\begin{tabular}{l|ccc|ccc|c}
\toprule
Text embedding & IR@1 & IR@5 & IR@10 & TR@1 & TR@5 & TR@10 & R$_\text{mean}$ \\
\midrule
\multicolumn{8}{l}{\textit{ResNet}} \\
\midrule
PDS & 14.8$\pm$0.2 & 38.3$\pm$0.2 & 50.4$\pm$0.2 & 22.6$\pm$0.6 & 44.9$\pm$0.5 & 56.9$\pm$0.6 & 38.0$\pm$0.2 \\
One inner step & 14.6$\pm$0.2 & 38.2$\pm$0.2 & 50.4$\pm$0.2 & 22.4$\pm$0.6 & 45.5$\pm$0.7 & 57.6$\pm$0.6 & 38.1$\pm$0.3 \\
Full unroll w/o $\Lanc$ & 14.5$\pm$0.2 & 38.6$\pm$0.3 & 51.8$\pm$0.4 & 21.7$\pm$0.5 & 45.5$\pm$0.7 & 57.6$\pm$0.6 & 38.3$\pm$0.1 \\
\cmidrule(lr){1-8}
Best caption & 12.1$\pm$0.2 & 32.5$\pm$0.6 & 45.2$\pm$0.2 & 16.5$\pm$0.3 & 40.7$\pm$0.4 & 52.3$\pm$0.4 & 33.2$\pm$0.2 \\
1-NN & 11.2$\pm$0.2 & 30.9$\pm$0.3 & 43.3$\pm$0.3 & 15.4$\pm$0.4 & 38.0$\pm$0.5 & 49.1$\pm$0.9 & 31.3$\pm$0.2 \\
5-NN & 14.0$\pm$0.2 & 36.6$\pm$0.2 & 49.3$\pm$0.3 & 20.0$\pm$0.5 & 42.6$\pm$0.8 & 54.7$\pm$0.7 & 36.2$\pm$0.3 \\
20-NN & 14.3$\pm$0.2 & 37.0$\pm$0.3 & 50.0$\pm$0.5 & 20.3$\pm$0.5 & 44.6$\pm$0.4 & 56.1$\pm$0.7 & 37.0$\pm$0.3 \\
Image embedding & 9.6$\pm$0.1 & 26.8$\pm$0.4 & 39.5$\pm$0.5 & 17.5$\pm$0.4 & 40.0$\pm$0.4 & 52.5$\pm$0.8 & 31.0$\pm$0.3 \\
Shift & 15.4$\pm$0.1 & 39.4$\pm$0.3 & 51.8$\pm$0.2 & 22.8$\pm$0.6 & 45.4$\pm$0.5 & 57.3$\pm$0.8 & 38.7$\pm$0.2 \\
\cmidrule(lr){1-8}
\cellcolor[gray]{0.93}TED & \cellcolor[gray]{0.93}16.4$\pm$0.2 & \cellcolor[gray]{0.93}41.8$\pm$0.2 & \cellcolor[gray]{0.93}54.9$\pm$0.1 & \cellcolor[gray]{0.93}24.3$\pm$1.0 & \cellcolor[gray]{0.93}49.5$\pm$0.3 & \cellcolor[gray]{0.93}62.8$\pm$0.5 & \cellcolor[gray]{0.93}41.6$\pm$0.2 \\
\midrule
\multicolumn{8}{l}{\textit{ViT}} \\
\midrule
PDS & 9.6$\pm$0.3 & 25.9$\pm$0.5 & 36.9$\pm$0.3 & 10.3$\pm$0.5 & 27.0$\pm$0.7 & 37.5$\pm$0.8 & 24.5$\pm$0.1 \\
One inner step & 9.7$\pm$0.3 & 26.0$\pm$0.5 & 36.9$\pm$0.2 & 10.5$\pm$0.5 & 27.8$\pm$0.8 & 38.1$\pm$0.4 & 24.8$\pm$0.1 \\
Full unroll w/o $\Lanc$ & 11.3$\pm$0.2 & 28.7$\pm$0.2 & 39.1$\pm$0.2 & 11.9$\pm$0.4 & 30.4$\pm$0.4 & 40.2$\pm$0.5 & 26.9$\pm$0.1 \\
\cmidrule(lr){1-8}
Best caption & 7.9$\pm$0.2 & 22.9$\pm$0.4 & 33.3$\pm$0.2 & 8.4$\pm$0.4 & 24.3$\pm$0.5 & 35.3$\pm$0.7 & 22.0$\pm$0.2 \\
1-NN & 7.3$\pm$0.4 & 21.3$\pm$0.3 & 31.2$\pm$0.4 & 9.6$\pm$0.5 & 23.7$\pm$0.8 & 32.5$\pm$0.7 & 20.9$\pm$0.3 \\
5-NN & 8.8$\pm$0.2 & 24.2$\pm$0.3 & 34.8$\pm$0.6 & 10.0$\pm$0.6 & 25.2$\pm$0.4 & 35.7$\pm$0.4 & 23.1$\pm$0.2 \\
20-NN & 8.4$\pm$0.1 & 23.9$\pm$0.4 & 34.8$\pm$0.3 & 10.4$\pm$0.8 & 27.1$\pm$0.3 & 37.5$\pm$0.6 & 23.7$\pm$0.3 \\
Image embedding & 6.4$\pm$0.3 & 19.1$\pm$0.4 & 28.2$\pm$0.5 & 8.7$\pm$0.6 & 22.7$\pm$1.1 & 32.0$\pm$1.3 & 19.5$\pm$0.5 \\
Shift & 9.5$\pm$0.1 & 26.0$\pm$0.2 & 37.1$\pm$0.3 & 10.4$\pm$0.5 & 27.5$\pm$0.6 & 38.2$\pm$0.5 & 24.8$\pm$0.2 \\
\cmidrule(lr){1-8}
\cellcolor[gray]{0.93}TED & \cellcolor[gray]{0.93}11.1$\pm$0.2 & \cellcolor[gray]{0.93}29.3$\pm$0.4 & \cellcolor[gray]{0.93}40.6$\pm$0.6 & \cellcolor[gray]{0.93}12.8$\pm$0.9 & \cellcolor[gray]{0.93}30.6$\pm$0.5 & \cellcolor[gray]{0.93}41.4$\pm$0.8 & \cellcolor[gray]{0.93}27.6$\pm$0.2 \\
\bottomrule
\end{tabular}
\end{table*}

\paragraph{All training-free text embeddings.}
\Cref{tab:supp_heuristics} shows that replacing $t^0$ by the best-matching caption, by the captions of the nearest real image or by the image embedding itself clearly lowers R$_\text{mean}$ at both budgets.
Averaging the captions of more neighbors reduces this loss, because the average moves back towards the cluster mean that $t^0$ already is.
Shift, which starts from $t^0$, is the only heuristic that improves PDS at both budgets, but its gain remains far below that of TED.

\begin{table*}[!t]
\centering
\footnotesize
\setlength{\aboverulesep}{0pt}
\setlength{\belowrulesep}{0pt}
\setlength{\extrarowheight}{2pt}
\caption{\textbf{Training-free text embeddings}. $\Delta$: gain over PDS averaged over the two architectures.}
\label{tab:supp_heuristics}
\begin{tabularx}{\textwidth}{l|YYY|YYY}
\toprule
 & \multicolumn{3}{c|}{$M{=}100$} & \multicolumn{3}{c}{$M{=}300$} \\
Text embedding & ResNet & ViT & $\Delta$ & ResNet & ViT & $\Delta$ \\
\midrule
PDS & 25.1$\pm$0.2 & 15.1$\pm$0.1 & -- & 38.0$\pm$0.2 & 24.5$\pm$0.1 & -- \\
Best caption & 19.8$\pm$0.2 & 13.3$\pm$0.2 & $-3.6$ & 33.2$\pm$0.2 & 22.0$\pm$0.2 & $-3.7$ \\
1-NN & 19.7$\pm$0.1 & 13.2$\pm$0.2 & $-3.7$ & 31.3$\pm$0.2 & 20.9$\pm$0.3 & $-5.1$ \\
5-NN & 23.7$\pm$0.2 & 15.2$\pm$0.2 & $-0.7$ & 36.2$\pm$0.3 & 23.1$\pm$0.2 & $-1.6$ \\
20-NN & 24.8$\pm$0.2 & 16.2$\pm$0.1 & $+0.4$ & 37.0$\pm$0.3 & 23.7$\pm$0.3 & $-0.9$ \\
Image embedding & 20.9$\pm$0.2 & 13.8$\pm$0.3 & $-2.8$ & 31.0$\pm$0.3 & 19.5$\pm$0.5 & $-6.0$ \\
Shift & 26.5$\pm$0.3 & 16.4$\pm$0.1 & $+1.3$ & 38.7$\pm$0.2 & 24.8$\pm$0.2 & $+0.5$ \\
\cellcolor[gray]{0.93}TED & \cellcolor[gray]{0.93}29.5$\pm$0.2 & \cellcolor[gray]{0.93}19.3$\pm$0.1 & \cellcolor[gray]{0.93}$+4.3$ & \cellcolor[gray]{0.93}41.6$\pm$0.2 & \cellcolor[gray]{0.93}27.6$\pm$0.2 & \cellcolor[gray]{0.93}$+3.4$ \\
\bottomrule
\end{tabularx}
\end{table*}

\paragraph{Unroll depth at $M{=}100$.}
\Cref{tab:supp_unroll} repeats the unroll-depth study at $M{=}100$, where a single step already covers all pairs because each epoch consists of one batch, and yet unrolling only the first step gives a small gain.
The gain increases steadily with the number of unrolled steps, in the same order as at $M{=}300$, and most of it is already reached after thirty steps, although the remaining steps still help on ViT, so that unrolling the full run is the best choice on average over the two architectures at both budgets.
A random window of $W$ consecutive steps, drawn anew at every outer step as in RaT-BPTT~\cite{feng2024embarrassingly}, back-propagates through as many steps as the first $K{=}W$ steps but gains less and moves the text less away from $t^0$, because each window covers only part of the training run and its gradient therefore misses the effect of the text on the rest of it.

\begin{table}[!tb]
\centering
\footnotesize
\setlength{\tabcolsep}{3pt}
\setlength{\aboverulesep}{0pt}
\setlength{\belowrulesep}{0pt}
\setlength{\extrarowheight}{2pt}
\caption{\textbf{Unroll depth}. Gain in R$_\text{mean}$ over PDS when the solver unrolls the first $K$ steps, or a random window of $W$ steps at every outer step.}
\label{tab:supp_unroll}
\begin{tabularx}{\columnwidth}{l|YYY|YYY}
\toprule
 & \multicolumn{3}{c|}{$M{=}100$} & \multicolumn{3}{c}{$M{=}300$} \\
Unroll & ResNet & ViT & Mean & ResNet & ViT & Mean \\
\midrule
first $K{=}1$ & +0.4 & +0.6 & +0.5 & +0.1 & +0.3 & +0.2 \\
first $K{=}3$ & +1.2 & +1.3 & +1.3 & +0.8 & +0.7 & +0.8 \\
first $K{=}10$ & +3.6 & +2.8 & +3.2 & +2.6 & +2.2 & +2.4 \\
first $K{=}30$ & +4.3 & +3.6 & +4.0 & +3.8 & +2.7 & +3.2 \\
full & +4.4 & +4.2 & +4.3 & +3.6 & +3.1 & +3.4 \\
\midrule
random $W{=}10$ & +2.3 & +1.7 & +2.0 & +1.9 & +1.6 & +1.8 \\
random $W{=}30$ & +2.9 & +2.4 & +2.6 & +2.6 & +1.8 & +2.2 \\
\bottomrule
\end{tabularx}
\end{table}

\begin{figure}[!tb]
\centering
\includegraphics[width=\linewidth]{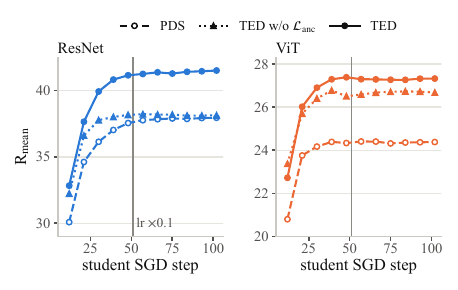}
\caption{\textbf{Student training dynamics}. Test R$_\text{mean}$ of ResNet and ViT students along training, averaged over five initializations, for PDS, TED without $\Lanc$ and TED. The vertical line marks the learning-rate decay.}
\label{fig:supp_dynamics}
\end{figure}

\paragraph{Training dynamics.}
\Cref{fig:supp_dynamics} tracks ResNet and ViT students during training on the PDS set, the TED set without $\Lanc$ and the TED set (Flickr30K, $M{=}300$).
After 12 steps, the ResNet TED student is $2.8$ points ahead of PDS, and the gap reaches $3.6$ by the end of training.
Without $\Lanc$, this student also starts ahead but ends only $0.3$ points above PDS.
The early lead comes from the unrolled objective, and $\Lanc$ is what preserves it for an encoder other than CLIP.

\paragraph{Anchor loss in every setting.}
\Cref{tab:supp_prior} compares TED without $\Lanc$ ($\lambda{=}0$) and with $\lambda{=}3$ in all PDS settings, and shows that the anchor improves the result in almost every setting, the only exception being ViT on Flickr30K at $M{=}100$, where the variant without $\Lanc$ is slightly better.
Without $\Lanc$, the gain on ResNet on Flickr30K drops sharply once the budget exceeds $M{=}100$, whereas with $\Lanc$ it remains large at every budget.
We attribute this to the larger number of pairs, which gives the optimization more freedom to fit the CLIP surrogate, so that the anchor becomes more important for $M\geq300$ than for $M{=}100$.

\begin{table}[!tb]
\centering
\footnotesize
\setlength{\tabcolsep}{3pt}
\setlength{\aboverulesep}{0pt}
\setlength{\belowrulesep}{0pt}
\setlength{\extrarowheight}{2pt}
\caption{\textbf{Anchor weight in every PDS setting}. Gain in R$_\text{mean}$ over PDS; the better value is in bold.}
\label{tab:supp_prior}
\begin{tabularx}{\columnwidth}{ll|YY|YY|YY}
\toprule
 & & \multicolumn{2}{c|}{$M{=}100$} & \multicolumn{2}{c|}{$M{=}300$} & \multicolumn{2}{c}{$M{=}500$} \\
Dataset & Backbone & $\lambda{=}0$ & $\lambda{=}3$ & $\lambda{=}0$ & $\lambda{=}3$ & $\lambda{=}0$ & $\lambda{=}3$ \\
\midrule
Flickr30K & ResNet & +3.4 & \textbf{+4.4} & +0.3 & \textbf{+3.6} & +1.5 & \textbf{+3.9} \\
Flickr30K & ViT & +4.5 & +4.2 & +2.4 & \textbf{+3.1} & +1.4 & \textbf{+3.1} \\
MS-COCO & ResNet & +0.9 & \textbf{+1.3} & +0.8 & \textbf{+1.7} & +1.3 & \textbf{+1.8} \\
MS-COCO & ViT & +1.1 & \textbf{+1.3} & +1.1 & \textbf{+1.8} & +1.1 & \textbf{+2.2} \\
\bottomrule
\end{tabularx}
\end{table}

\paragraph{Starting point and anchor.}
\Cref{tab:supp_init} starts the optimization from random directions that are nearly orthogonal to $t^0$ instead of from $t^0$ itself (Flickr30K, $M{=}300$).
With $\Lanc$ centered on $t^0$, the random start recovers almost the full gain of TED and ends close to $t^0$, whereas without $\Lanc$ it reaches a held-out surrogate loss close to that of the $t^0$ start but brings almost no gain on ResNet.
Minimizing the surrogate loss thus leads to equally good solutions with or without $t^0$, and only $\Lanc$ selects the one that transfers, so $t^0$ matters as the center of $\Lanc$ rather than as the starting point.

\begin{table}[!tb]
\centering
\footnotesize
\setlength{\tabcolsep}{3pt}
\setlength{\aboverulesep}{0pt}
\setlength{\belowrulesep}{0pt}
\setlength{\extrarowheight}{2pt}
\caption{\textbf{Starting point and anchor loss}. Held-out $J$: surrogate loss on held-out training images at the selected step; R$_\text{mean}$ (\%), mean $\pm$ std over five student initializations.}
\label{tab:supp_init}
\begin{tabularx}{\columnwidth}{cc|YY|YY}
\toprule
Start & $\Lanc$ & $\cos(t^\star, t^0)$ & Held-out $J$ & ResNet & ViT \\
\midrule
PDS ($t^0$) & -- & 1.000 & -- & 38.0$\pm$0.2 & 24.5$\pm$0.1 \\
$t^0$ & \xmark & 0.695 & 4.01 & 38.3$\pm$0.1 & 26.9$\pm$0.1 \\
\cellcolor[gray]{0.93}$t^0$ & \cellcolor[gray]{0.93}\cmark & \cellcolor[gray]{0.93}0.956 & \cellcolor[gray]{0.93}4.17 & \cellcolor[gray]{0.93}41.6$\pm$0.2 & \cellcolor[gray]{0.93}27.6$\pm$0.2 \\
random & \xmark & 0.347 & 4.09 & 38.2$\pm$0.2 & 26.3$\pm$0.3 \\
random & \cmark & 0.928 & 4.18 & 41.2$\pm$0.4 & 27.6$\pm$0.4 \\
\bottomrule
\end{tabularx}
\end{table}

\paragraph{Trajectory-matching objective.}
\Cref{tab:supp_mtt} replaces $J$ by the trajectory-matching loss of MTT-VL and LoRS and keeps everything else of TED (\cref{sec:supp_sets}), and shows that, with $\Lanc$, the configuration selected on the validation split improves PDS but remains below TED on both architectures.
Without $\Lanc$, trajectory matching moves the text far from $t^0$ and brings only a small gain, although its held-out surrogate loss is close to that of TED, which is the same pattern as for TED without $\Lanc$.
On the validation split, the gain grows with the length of the matched segment from $N{=}8$ to $N{=}50$ student steps, so the closer the objective comes to the full training run that TED unrolls, the better the text transfers, and trajectory matching in addition requires expert trajectories on the full training set, which TED does not.

\paragraph{Closed-form parameter matching.}
This variant keeps the images, the surrogate, the parametrization of \cref{eq:supp_param} and the anchor loss of TED, and replaces $J$ by the analytic parameter matching loss of APM~\cite{bo2026apm}.
For row-stacked features $A$ and $B$ of $n$ pairs, let $\Sigma_{AB}=\frac{1}{n}(A-\mathbf{1}\mu_A^\top)^\top(B-\mathbf{1}\mu_B^\top)$, with $\mu_A$ and $\mu_B$ the feature means.
Since the student of the evaluation protocol has no image projection, the image features serve as the image-side representation, and a teacher head $g_{\bar\theta}$ with the architecture of $g_\theta$, trained for ten epochs on the real training pairs in the surrogate space, provides the text-side representation.
For image features $E$, text embeddings $T$ and teacher outputs $V=g_{\bar\theta}(T)$, the closed-form parameters of the text and image projections are
\begin{equation}
\begin{aligned}
  W_T &= (\Sigma_{TT}+\alpha I)^{-1}\,\Sigma_{TE}\,(\Sigma_{EE}+\alpha I)^{-1},\\
  W_I &= (\Sigma_{EE}+\alpha I)^{-1}\,\Sigma_{EV}\,(\Sigma_{VV}+\alpha I)^{-1},
\end{aligned}
\end{equation}
with the regularization $\alpha{=}0.05$ of APM.
The real parameters $W^\ast_T$ and $W^\ast_I$ are computed once from $f_I(x_n)$ and $f_T(y_n)$, with every image paired with its $k$-th caption for $k=1,\dots,5$ as in APM, and the synthetic parameters $\widetilde W_T$ and $\widetilde W_I$ from $\{(e_m,t_m)\}_{m=1}^M$.
The solver minimizes
\begin{equation}
  \mathcal{L}_{\text{APM}}(T)=\frac{\|W^\ast_I-\widetilde W_I\|_F^2}{\|W^\ast_I\|_F^2}+\frac{\|W^\ast_T-\widetilde W_T\|_F^2}{\|W^\ast_T\|_F^2}+\lambda\,\Lanc(U),
\end{equation}
cycling over the five caption subsets across outer steps, with $\lambda{=}3$ and the optimizer and 300 outer steps of TED.
The normalization by the real parameters places the matching loss on the scale of $J$ at $t^0$, so that $\lambda$ weighs the anchor loss as in TED.

\begin{table}[!tb]
\centering
\footnotesize
\setlength{\tabcolsep}{3pt}
\setlength{\aboverulesep}{0pt}
\setlength{\belowrulesep}{0pt}
\setlength{\extrarowheight}{2pt}
\caption{\textbf{Objective}. Trajectory matching replaces $J$ and keeps the images, surrogate, parametrization and anchor loss of TED; its configuration is selected on the validation split with and without $\Lanc$. R$_\text{mean}$ (\%), mean $\pm$ std over five student initializations; $\Delta$: gain over PDS averaged over the two architectures.}
\label{tab:supp_mtt}
\begin{tabularx}{\columnwidth}{lc|YYY}
\toprule
Objective & $\Lanc$ & ResNet & ViT & $\Delta$ \\
\midrule
PDS ($t^0$) & -- & 38.0$\pm$0.2 & 24.5$\pm$0.1 & -- \\
Trajectory matching & \xmark & 38.2$\pm$0.1 & 25.5$\pm$0.4 & $+0.6$ \\
Trajectory matching & \cmark & 40.3$\pm$0.2 & 26.6$\pm$0.2 & $+2.2$ \\
APM-style matching & \xmark & 35.3$\pm$0.3 & 23.7$\pm$0.2 & $-1.8$ \\
APM-style matching & \cmark & 37.3$\pm$0.2 & 24.8$\pm$0.1 & $-0.2$\\
\cellcolor[gray]{0.93}TED & \cellcolor[gray]{0.93}\cmark & \cellcolor[gray]{0.93}41.6$\pm$0.2 & \cellcolor[gray]{0.93}27.6$\pm$0.2 & \cellcolor[gray]{0.93}$+3.4$ \\
\bottomrule
\end{tabularx}
\end{table}

\section{Analysis of the Solved Embeddings}
\label{sec:supp_analysis}

\paragraph{Distinctiveness across synthetic sets.}
\Cref{tab:supp_distinct} reports the within-set distinctiveness of PDS (Flickr30K), EDGE (Flickr30K) and PDS (MS-COCO), all with $M{=}300$.
On all three sets TED raises the margin between each text's own image and the hardest other image and nearly removes the texts that prefer another image, whereas a random move of the same angle and a permutation of the moves across images leave both unchanged or worse.
The margin rises by about $0.04$ on every set, and the number of corrected texts is largest for PDS on MS-COCO, whose cluster texts are the least distinctive to begin with: 86 of 300 texts prefer another image under $t^0$ and 10 under TED.

\begin{table}[!tb]
\centering
\caption{\textbf{Within-set distinctiveness}. Margin: mean of own-image cosine minus the highest other-image cosine; $<0$: texts that prefer another image; R@1: text-to-image retrieval within the set (\%). Random and permuted: same angle as TED, mean over three seeds.}
\label{tab:supp_distinct}
\scriptsize
\setlength{\tabcolsep}{1.2pt}
\setlength{\aboverulesep}{0pt}
\setlength{\belowrulesep}{0pt}
\setlength{\extrarowheight}{2pt}
\begin{tabularx}{\columnwidth}{l|YYY|YYY|YYY}
\toprule
 & \multicolumn{3}{c|}{PDS Flickr30K} & \multicolumn{3}{c|}{EDGE Flickr30K} & \multicolumn{3}{c}{PDS MS-COCO} \\
Text & Margin & $<0$ & R@1 & Margin & $<0$ & R@1 & Margin & $<0$ & R@1 \\
\midrule
$t^0$ & 0.030 & 56 & 81.3 & 0.061 & 13 & 95.7 & 0.016 & 86 & 71.3 \\
Random & 0.028 & 60 & 80.0 & 0.056 & 16 & 94.8 & 0.015 & 87 & 70.9 \\
Permuted & 0.024 & 67 & 77.6 & 0.051 & 21 & 93.1 & 0.013 & 90 & 70.1 \\
\rowcolor[gray]{0.93}TED & \textbf{0.071} & \textbf{5} & \textbf{98.3} & \textbf{0.100} & \textbf{1} & \textbf{99.7} & \textbf{0.056} & \textbf{10} & \textbf{96.7} \\
\bottomrule
\end{tabularx}
\end{table}

\begin{table}[!tb]
\centering
\footnotesize
\setlength{\aboverulesep}{0pt}
\setlength{\belowrulesep}{0pt}
\setlength{\extrarowheight}{2pt}
\caption{\textbf{TED on real images}. Generated: the PDS images (Flickr30K, $M{=}100$). Nearest real: for each of their image prototypes, a real training image nearest to it in CLIP space. Both image sets use the same PDS text embeddings $t^0$.}
\label{tab:realimg}
\begin{tabularx}{\columnwidth}{ll|YY}
\toprule
Images & Text & ResNet & ViT \\
\midrule
\multirow{2}{*}{Generated} & PDS & 25.1$\pm$0.2 & 15.1$\pm$0.1 \\
 & \cellcolor[gray]{0.93}TED & \cellcolor[gray]{0.93}\textbf{29.5}$\pm$0.2 & \cellcolor[gray]{0.93}\textbf{19.3}$\pm$0.1 \\
\midrule
\multirow{2}{*}{Nearest real} & PDS & 19.5$\pm$0.2 & 12.3$\pm$0.0 \\
 & \cellcolor[gray]{0.93}TED & \cellcolor[gray]{0.93}25.8$\pm$0.2 & \cellcolor[gray]{0.93}16.5$\pm$0.2 \\
\bottomrule
\end{tabularx}
\end{table}

\paragraph{Real images.}
Since TED only requires image features and initial text embeddings, it also applies to real images. In \cref{tab:realimg}, we replace each image of the PDS set (Flickr30K, $M{=}100$) by a real training image nearest to its prototype and keep the text embeddings unchanged.
TED improves the real images on both architectures.
The PDS text describes the center of a cluster, which an image generated from that center matches more closely than its nearest real neighbor, so the original text fits the real images less well and leaves more for TED to correct.
After TED, the real images still trail the generated ones on both architectures, and we attribute the remaining gap to the images themselves, which the text cannot change.

\FloatBarrier
\section{Qualitative Results}
\label{sec:supp_qualitative}

\paragraph{What the distilled text encodes.}
\begin{figure*}[!t]
\centering
\includegraphics[width=\textwidth]{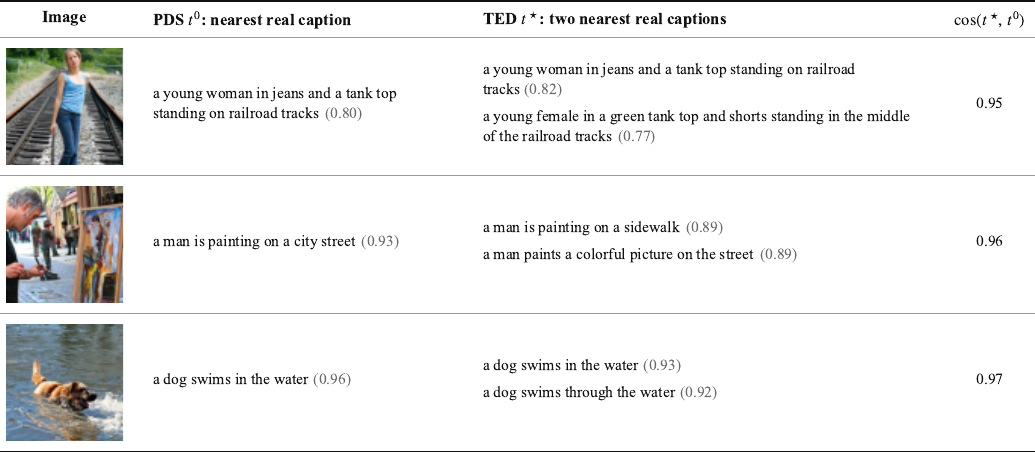}
\caption{\textbf{Distilled text embeddings}. Synthetic images at the 25th, 50th and 75th percentiles of $\cos(t^\star, t^0)$, each with the nearest real training caption of the PDS embedding $t^0$ and the two nearest distinct captions of the TED embedding $t^\star$. CLIP-L cosines in brackets.}
\label{fig:qualitative}
\end{figure*}

\Cref{fig:qualitative} maps the solved text embeddings back to their nearest real training captions (PDS, Flickr30K, $M{=}300$). The mean $\cos(t^\star, t^0)$ is 0.956 and 63\% of the images keep the same nearest caption, so TED keeps the subject: the man still paints in the street and the dog still swims. What changes is where the embedding sits among the captions. The mean cosine to the nearest caption falls from 0.897 to 0.869, and in 97\% of the images the nearest caption of $t^\star$ is less similar than that of $t^0$. On average, $t^\star$ is closer to the mean of its five nearest captions (0.917) than to its nearest single caption, and the painter's $t^\star$ is equally close (0.89) to ``a man is painting on a sidewalk'' and ``a man paints a colorful picture on the street''. TED therefore writes a text target that lies between several real phrasings of the scene rather than on any one of them.

\begin{figure*}[!t]
\centering
\includegraphics[width=\textwidth]{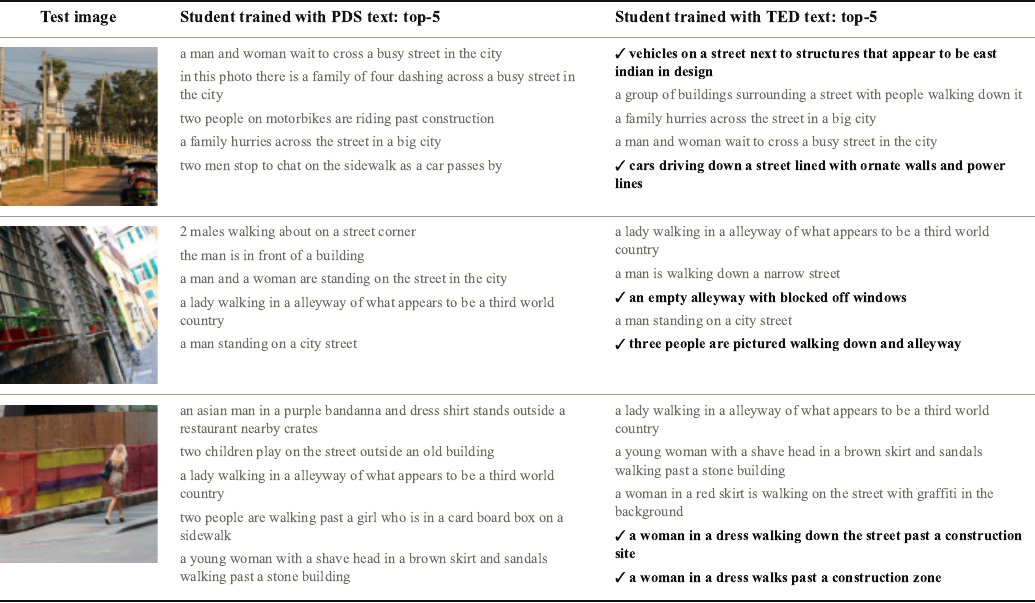}
\caption{\textbf{Retrieval examples}. Test images with no ground truth in the top five of the PDS student, ranked by the number of ground truths in the top five of the TED student. Ground-truth captions in bold with a check mark.}
\label{fig:supp_retrieval_rule}
\end{figure*}

\begin{figure*}[!t]
\centering
\includegraphics[width=\textwidth]{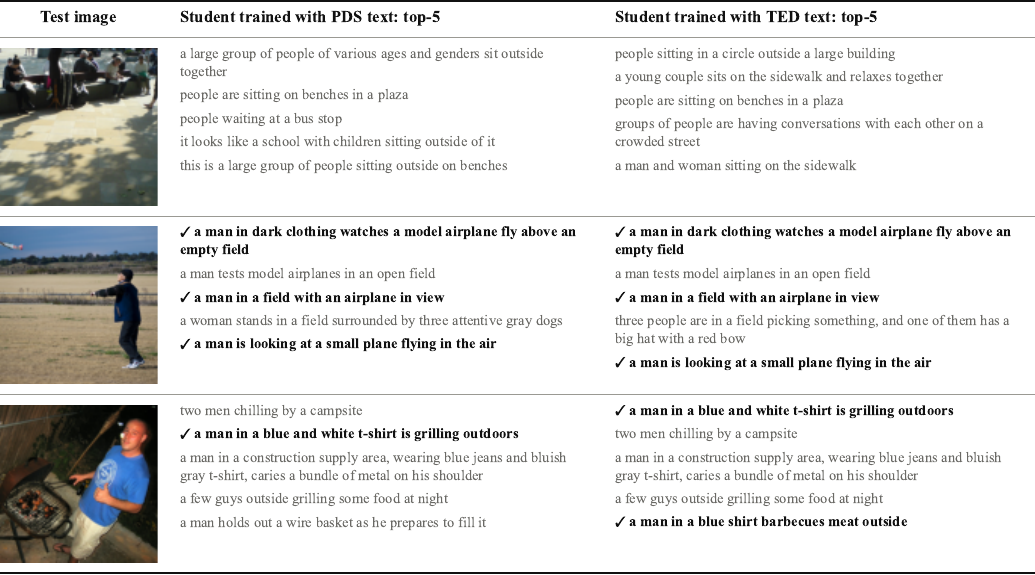}
\caption{\textbf{Retrieval examples, including failure cases}. Same setting as \cref{fig:supp_retrieval_rule}.}
\label{fig:supp_retrieval_random}
\end{figure*}

\paragraph{Retrieval examples.}
\Cref{fig:supp_retrieval_rule,fig:supp_retrieval_random} query Flickr30K test images against all $5{,}000$ test captions with two ResNet students trained on the same $500$ PDS images with the same seed, one on the PDS text and one on the TED text.
Over the $1{,}000$ test images, the number with a ground-truth caption in the top five rises from $486$ to $555$ and the number with a correct first result from $226$ to $266$; TED turns a miss into a hit for $109$ images and a hit into a miss for $40$.
In the selected examples, the PDS student returns captions of other street scenes for a temple road, an alley and a construction fence, whereas the TED student retrieves captions that name the specific scene: the ornate walls and power lines, the blocked-off windows, the construction site.
The same images train both students, so these differences come from the text alone.

\begin{figure*}[p]
\centering
\includegraphics[width=\textwidth]{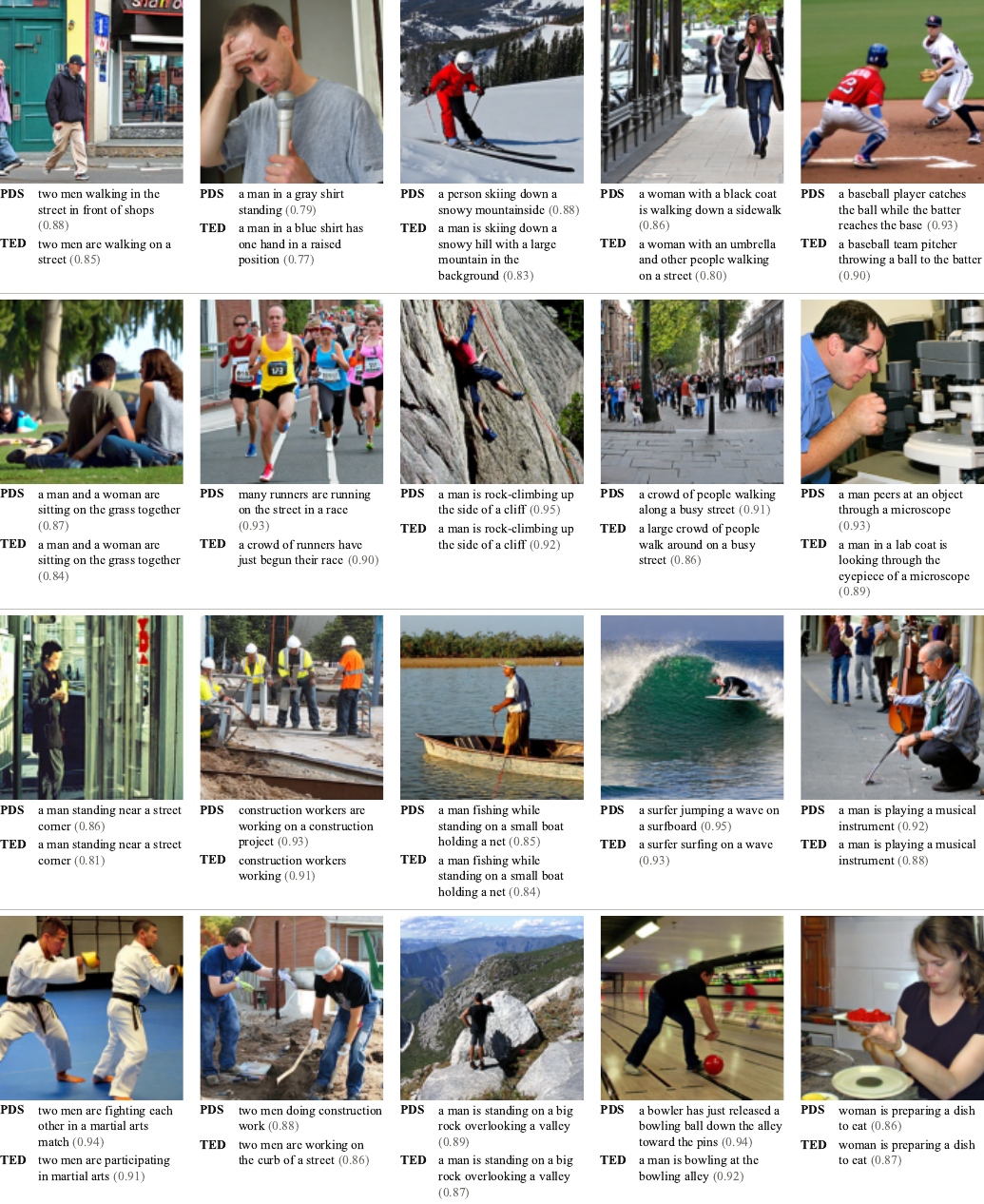}
\caption{\textbf{PDS synthetic images with PDS and TED text}. Nearest real training caption of $t^0$ (PDS) and of $t^\star$ (TED) for each image, with CLIP-L cosines.}
\label{fig:supp_gallery_flickr}
\end{figure*}

\begin{figure*}[p]
\centering
\includegraphics[width=\textwidth]{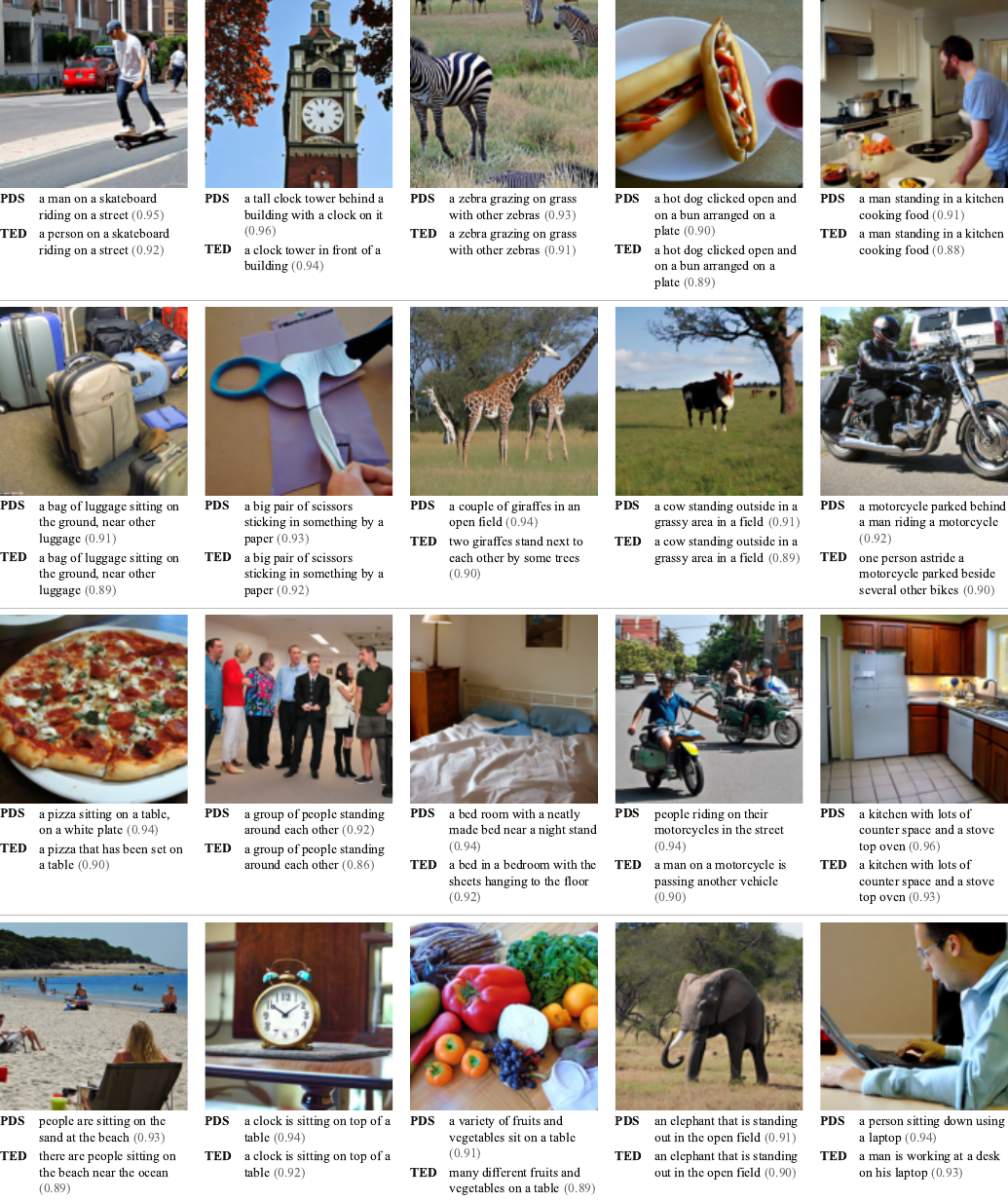}
\caption{\textbf{PDS synthetic images with PDS and TED text}. Same layout and selection as \cref{fig:supp_gallery_flickr}.}
\label{fig:supp_gallery_coco}
\end{figure*}

\paragraph{PDS synthetic sets with distilled text.}
\Cref{fig:supp_gallery_flickr,fig:supp_gallery_coco} show 20 PDS images of each dataset at $M{=}100$, each with the nearest real training caption of $t^0$ and of $t^\star$.
On both datasets, $t^\star$ keeps the nearest caption of $t^0$ for nearly half of the images, and for most images it lies slightly farther from its nearest caption than $t^0$ does.
Where the nearest caption changes, the new caption keeps the scene and its subject and only rephrases it, for example ``a man peers at an object through a microscope'' becomes ``a man in a lab coat is looking through the eyepiece of a microscope'', and ``a couple of giraffes in an open field'' becomes ``two giraffes stand next to each other by some trees''.
As at $M{=}300$ (\cref{fig:qualitative}), TED thus places each text between several real descriptions of its image rather than on any single one of them.

\section{Limitations and Broader Impact}
\label{sec:supp_limitations}

TED solves for text embeddings in the space of one text encoder, CLIP-L or BERT, and its output is an embedding rather than a sentence, so students that read raw text cannot use it.
Its gains persist when the image encoder is trained (\cref{sec:supp_beyond}), but depend on the training protocol: at $M{=}500$ on NFNet, TED improves LoRS by $2.0$ points under the same training and lowers it by $4.0$ under LoRS's native training.
The method changes only the text side, so the quality of a distilled set remains bounded by its images.
Distilled sets reduce the compute and storage needed to train retrieval models; they also inherit the content and biases of the source dataset and of the generator, and a smaller set does not remove them.

\end{document}